\documentclass[lettersize,journal]{IEEEtran}
\usepackage{amsmath,amsfonts}
\usepackage[ruled,vlined]{algorithm2e}
\usepackage{amsfonts}     
\usepackage{pifont}
\usepackage{array}
\usepackage[caption=false,font=normalsize,labelfont=sf,textfont=sf]{subfig}
\usepackage{textcomp}
\usepackage{stfloats}
\usepackage{url}
\usepackage{verbatim}
\usepackage{graphicx}
\usepackage{cite}
\usepackage{microtype}  
\usepackage{xcolor, array, colortbl}

\newcommand{\mf}{\mathbf}
\newcommand{\mb}{\mathbb}
\newcommand{\mr}{\mathrm}
\newcommand{\mc}{\mathcal}

\newcommand{\xmarkg}{\textcolor{gray}{\ding{55}}\xspace}%
\newcommand{\cmark}{\ding{51}}

\usepackage{amsthm}
 
\usepackage{multicol}

\usepackage{psfrag}
\definecolor{deepred}{rgb}{0.698,0.133,0.133}
\definecolor{blue}{rgb}{0,0,1}
\definecolor{lightgray}{rgb}{.93,.93,.93}
\definecolor{blue}{rgb}{0,0,1}
\usepackage{booktabs}
\usepackage{wrapfig}

\usepackage{makecell}	
\usepackage{fancyhdr}
\usepackage{multirow}
\usepackage{rotating}
\usepackage{color}

\usepackage[pagebackref=false,breaklinks=true,colorlinks,bookmarks=false,linkcolor=red,urlcolor=blue,citecolor=blue]{hyperref}

\begin{document}

\title{Never-Ending Behavior-Cloning Agent for Robotic Manipulation}

\author{%
    Wenqi Liang, Gan Sun,~\IEEEmembership{Member,~IEEE}, Yao He, Yu Ren, Jiahua Dong and Yang Cong
  
  \thanks{Wenqi Liang and Yu Ren are with the State Key Laboratory of Robotics, Shenyang Institute of Automation, Chinese Academy of Sciences, Shenyang 110016, China, and also with the University of Chinese Academy of Sciences, Beijing 100049 China. Email: liangwenqi0123@gmail.com, renyu0414@gmail.com.}
\thanks{Yao He, Gan Sun and Yang Cong are with the School of Automation Science and Engineering, South China University of Technology, Guangzhou, 510640, China. Email: sungan1412@gmail.com,  heyao0293@gmail.com, congyang81@gmail.com.}
\thanks{Jiahua Dong is with the Mohamed bin Zayed University of Artificial Intelligence, Abu Dhabi, United Arab Emirates. Email: dongjiahua1995@gmail.com.}
\thanks{This work is supported by the National Nature Science Foundation of China under Grant (62273333, 62225310, 62127807),  and the Fundamental Research Funds for the Central Universities (2024ZYGXZR024, 2025ZYGXZR032).}
\thanks{The corresponding author is \emph{Prof. Gan Sun.}}}

\markboth{Journal of \LaTeX\ Class Files,~Vol.~14, No.~8, August~2021}%
{Shell \MakeLowercase{\textit{et al.}}: A Sample Article Using IEEEtran.cls for IEEE Journals}


\maketitle

\begin{abstract}
Relying on multi-modal observations, embodied robots (\emph{e.g.,} humanoid robots) could perform multiple robotic manipulation tasks in unstructured real-world environments. However, most language-conditioned behavior-cloning agents in robots still face existing long-standing challenges, \emph{i.e.,} 3D scene representation and human-level task learning, when adapting into a series of new tasks in practical scenarios. We here investigate these above challenges with NBAgent in embodied robots, a pioneering language-conditioned \underline{N}ever-ending \underline{B}ehavior-cloning \underline{Agent}, which can continually learn observation knowledge of novel 3D scene semantics and robot manipulation skills from skill-shared and skill-specific attributes, respectively. Specifically, we propose a skill-shared semantic rendering module and a skill-shared representation distillation module to effectively learn 3D scene semantics from skill-shared attribute, further tackling 3D scene representation overlooking. Meanwhile, we establish a skill-specific evolving planner to perform manipulation knowledge decoupling, which can continually embed novel skill-specific knowledge like human from latent and low-rank space. Finally, we design a never-ending embodied robot manipulation benchmark, and expensive experiments demonstrate the significant performance of our method.

\end{abstract}


\section{Introduction}
\IEEEPARstart{R}{obot} learning has attracted growing interest in integrating machine learning with a robot control system to solve various robotic tasks, such as manipulation \cite{shridhar2023perceiver}, navigation \cite{schumann2024velma}, mapping and localization \cite{lajoie2023swarm}. In recent decades, behavior-cloning agents \cite{goyal2023rvt, ze2023gnfactor} have demonstrated notable success in effective training with only a few demonstrations and direct implementation in real robots. Benefiting from vision-language observations, current multi-task behavior-cloning methods focus on utilizing multi-modal data to efficiently execute complex manipulation tasks with visual observations. For instance, PerAct~\cite{shridhar2023perceiver} utilizes a PerceiverIO Transformer \cite{jaegle2021perceiver} to encode language goals and RGB-D voxel observations, subsequently generating discretized robotic actions. 3D Diffuser Actor \cite{ke20243d} introduces a novel 3D denoising transformer that integrates 3D visual scene representations, language instructions, and proprioception to predict noise in 3D robot pose trajectories.

\begin{figure}[t]
\centering
\vspace{-3pt}
\includegraphics[width = 1.0\linewidth]
{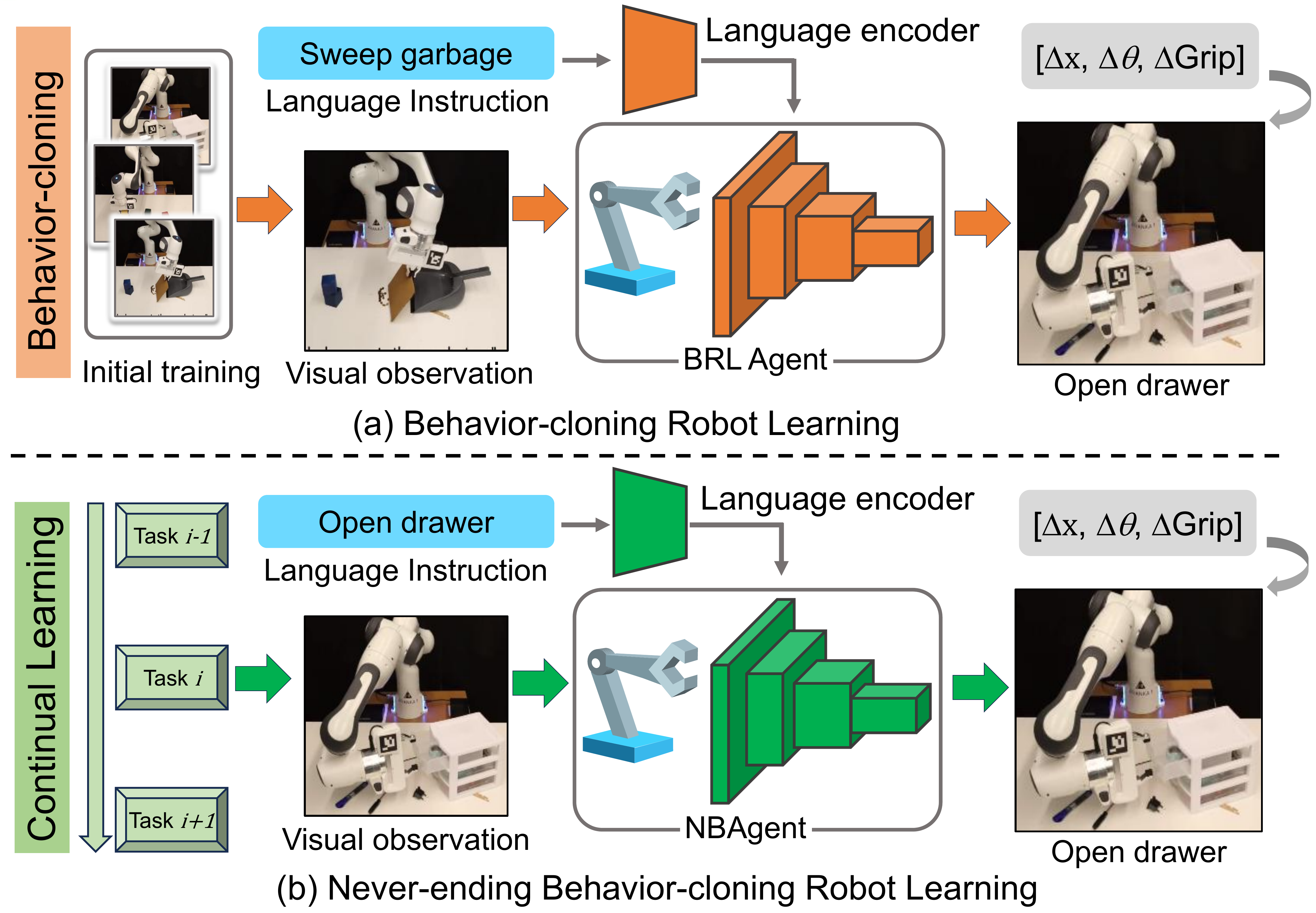}
\vspace{-21pt}
\caption{Demonstration illustration of our proposed never-ending behavior-cloning robot learning. As illustrated in (a), behavior-cloning robot learning primarily focuses on initially training on a fixed dataset, subsequently relying on the generalization capability to execute tasks in unseen environments, where a pre-trained CLIP model \cite{radford2021learning} serves as the the language encoder to process the language instruction. As depicted in (b), the never-ending behavior-cloning framework enables robotic systems to progressively acquire novel manipulation skills in a continual learning manner, thereby demonstrating enhanced adaptability and generalization capabilities when confronted with unseen and challenging tasks. }
\label{fig: motivation}
\vspace{-4pt}
\end{figure}

However, most existing language-conditioned behavior-cloning methods \cite{jaegle2021perceiver, goyal2023rvt, ke20243d} assume that the skills mastered by a robot remain unchanged over time in real-world 2D scenarios, and focus on training policy model on a fixed set of manipulation tasks. As shown in Fig.~\ref{fig: motivation}(a), they are generalization-constrained when undertaking new skills and handling novel objects with intricate structures \cite{wan2023lotus}. To address this scenario, the embodied robots—\emph{like human}—should be capable of learning novel challenging manipulation skills in a continual learning manner. For instance, a home robot in the open-ended world is expected to consecutively learn various novel manipulation skills with demonstrations of behavior, and timely meet the evolving needs of its owners. Generally, a trivial approach for this scenario involves retraining the robot on all past and novel data, which leads to a large computational burden and high cost of memory storage, thereby is limited in real-world unstructured 3D applications. 

To tackle the aforementioned challenges in real-world robotic applications, we formulate and introduce a practically significant problem framework, \emph{i.e.}, \underline{N}ever-ending \underline{B}ehavior-cloning \underline{R}obot \underline{L}earning (NBRL).
As conceptually depicted in Fig.~\ref{fig: motivation}(b), the proposed NBRL framework empowers embodied agents to engage in never-ending skill acquisition through continual learning, while  effectively mitigating catastrophic forgetting on learned skills. Different from learning category-wise knowledge focused by most existing methods \cite{rebuffi2017icarl, wan2023lotus, dong2023heterogeneous}, skill-wise knowledge encompasses object understanding, scene understanding, action semantics, language grounding, and operation sequences. We here take the attempt to rethink some attributes about skill-wise knowledge: 

\textbullet ~ \textbf{Skill-Shared Attribute} indicates that different complex skills share common knowledge in object understanding, scene understanding, action semantics, language grounding, and operation sequences. For example, the action semantics of \textit{pouring} and the scene understanding required in both  \textit{pouring water into a cup} and \textit{pouring juice into a bowl} illustrate this shared knowledge. We primarily investigate the consistency of skill-shared knowledge across various skills on semantics of 3D scenes, which plays a key role in addressing skill forgetting and achieving comprehensive understanding on 3D scenes. 

\textbullet ~ \textbf{Skill-Specific Attribute} denotes the unique knowledge components associated with different skills, including object understanding, scene understanding, action semantics, language grounding, and operation sequences, \emph{e.g.}, distinguishing the object understanding of a \textit{cup} versus a \textit{bowl}, or task-specific variations in operation sequences.
Focusing the embodied robot to learn skill-specific knowledge can effectively perform novel skills learning and tackle forgetting on past learned manipulation skills or tasks.

To tackle the above-mentioned challenges, we propose a pioneering language-conditioned \underline{N}ever-ending \underline{B}ehavior-cloning \underline{Agent} (\emph{i.e.,} NBAgent), which can continually acquire skill-wise knowledge from both skill-shared and skill-specific attributes. To the best of our knowledge, this is an earlier attempt to explore continual learning for multi-modal behavior-cloning robotic manipulation. To be specific, we design a skill-shared semantic rendering module (SSR) and a skill-shared representation distillation module (SRD) to transfer skill-shared knowledge on semantics of 3D scenes. Supervised by Neural Radiance Fields (NeRFs) and a vision foundation model, SSR can transfer skill-shared semantic from 2D space into 3D space across novel and old skills. Meanwhile, SRD can effectively distill skill-shared knowledge between old and current models to well align voxel representation. Additionally, we propose a skill-specific evolving planner (SEP) to decouple the skill-wise knowledge in a latent and low-rank space, and focus on skill-specific knowledge learning to alleviate manipulation skill forgetting. Extensive experiments on robotic manipulation tasks validate that our proposed NBAgent model achieves state-of-the-art performance in NBRL problem.

Several major contributions of our work are as follows:
\begin{itemize}
\item  We take the earlier attempt to explore a practical challenging problem called {N}ever-ending Behavior-cloning Robot Learning (NBRL), where we propose \underline{N}ever-ending \underline{B}ehavior-cloning \underline{Agent} (\emph{i.e.,} NBAgent) to address the core challenges of skill-wise knowledge learning in 3D scene from skill-shared and skill-specific attributes.

\item  We develop a skill-shared semantic rendering module and a skill-shared representation distillation module to capture the 3D semantic knowledge, which can transfer skill-shared semantic of 3D scenes to overcome 3D reasoning overlooking in continual skill learning. 


\item  We design a skill-specific evolving planner to perform skill-specific knowledge learning, which can decouple the skill-wise knowledge into a latent and low-rank space, and continually embed novel skill-specific knowledge. Comprehensive experiments conducted on our proposed NBRL benchmark for home robotic manipulation demonstrate the effectiveness and robustness of our method.
\end{itemize}

\begin{figure*}[t]
\centering
\includegraphics[width = 1.0\linewidth]
{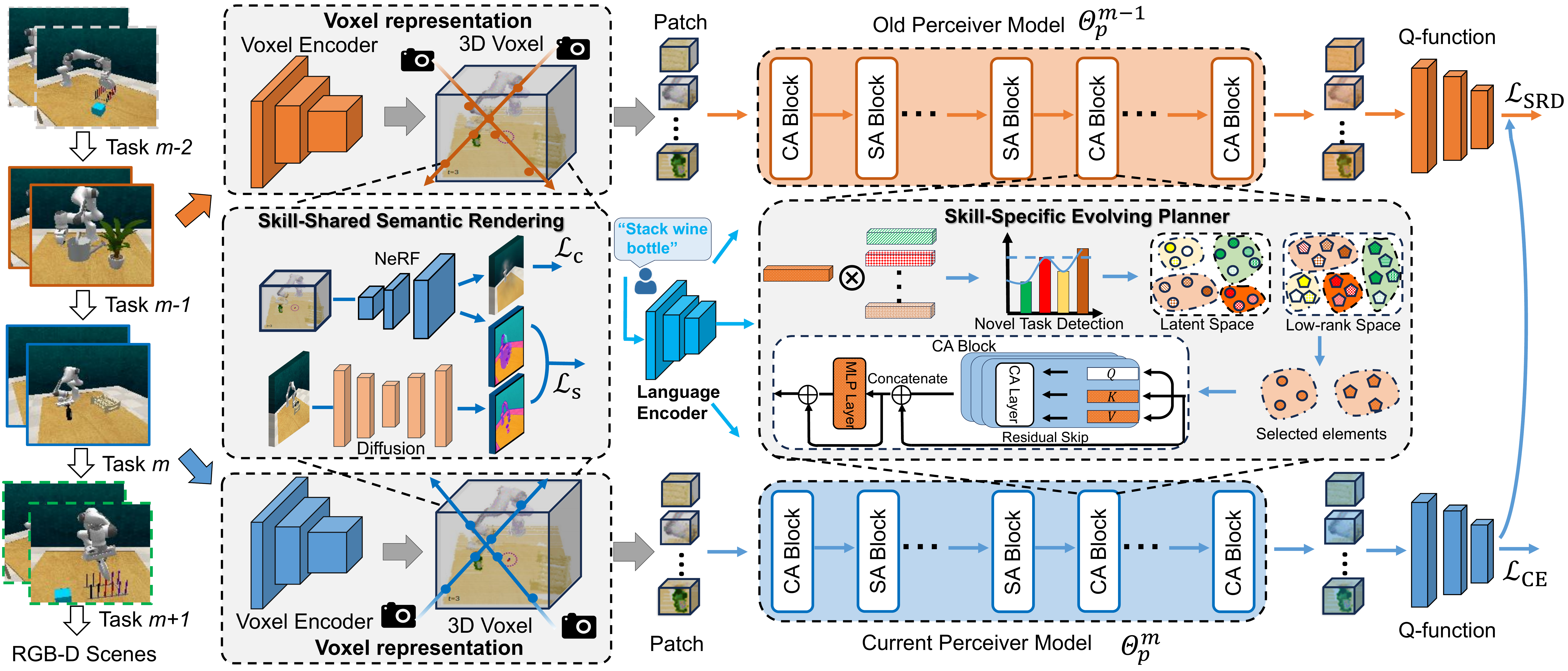}
\vspace{-20pt}
\caption{Overview of the proposed NBAgent. It consists of a \textit{skill-shared semantic rendering module} and a \textit{skill-shared representation distillation loss} $\mc{L}_{\mr{SRD}}$ to transfer skill-shared knowledge on  semantics of 3D scenes and overcome 3D reasoning overlooking
in continual learning, and a \textit{skill-specific evolving planner} to learn skill-specific knowledge, addressing catastrophic forgetting on learned skills.}
\label{fig: overview}
\end{figure*}

\section{Related Work}
\textbf{Robotic Manipulation.}
Recent works \cite{brohan2023can,liang2025pixelvla,chowdhery2023palm,huang2022inner,shah2023lm,driess2023palm,brohan2022rt,zitkovich2023rt} have resulted in substantial advancements in robotic manipulation utilizing mult-imodal observation. Generalizable observation representation and language understanding are the key factors in achieving effective task reasoning and robust policy execution. VIMA \cite{jiang2022vima} proposes a novel multi-modal prompting scheme, which transforms a wide range of robotic manipulation tasks into a sequence modeling problem. Octo\cite{team2024octo} enhances robotic manipulation with a transformer-based policy, supporting versatile task definitions and efficient fine-tuning. GPT-4V(ision)\cite{wake2024gpt} enables robots to perform tasks from human demonstrations by generating executable programs from video inputs. Compared with directly using images as the manipulation input \cite{goyal2022ifor,shridhar2022cliport}, voxelizing 3D point clouds as a 3D representation \cite{shridhar2023perceiver,ze2023gnfactor,goyal2023rvt,james2022coarse} can accomplish more complex tasks. PPM6D \cite{yu20246} performs differentiable 6-D pose estimation for robotic grasping by cross-fusing RGB–point-cloud features with attention-guided point-pair matching, while \cite{liu2022distributed} develops a consensus-driven distributed off-policy actor–critic algorithm with Lyapunov-based convergence guarantees and validates it on multi-UR5 robotic arm control. GNFactor \cite{ze2023gnfactor} can generalize neural radiance field rendering to enhance 3D representation through joint training, improving the model’s generalization ability. ManiGaussian\cite{lu2024manigaussian} enhances robotic manipulation using dynamic Gaussian Splatting to model spatiotemporal dynamics and predict optimal actions via future scene reconstruction. 3D Diffuser Actor\cite{ke20243d} combines 3D scene representations with diffusion models to achieve SOTA performance in robotic manipulation.


\textbf{Continual Learning.}
Continual learning provides the foundation for the adaptive development of AI systems \cite{wang2023comprehensive, zhao2023adaptcl, dong2024continually}. The main approaches of continual learning can be categorized into three directions: Parameter
regularization-based methods \cite{rebuffi2017icarl,li2017learning,derakhshani2021kernel,douillard2020podnet,dong2023heterogeneous} balance the old and new tasks by adding more explicit regularization terms. C-LoRA\cite{smith2024continualdiffusion} reduces catastrophic forgetting in diffusion models via self-regularized low-rank adaptation. RTRA\cite{nokhwal2023rtra} accelerates continual learning training with natural gradient descent, improving efficiency while maintaining effective performance.
 Architecture-based methods \cite{jung2020continual,wu2021incremental,wang2022coscl,toldo2022bring} construct network parameters for different tasks. AdaB2N \cite{lyu2024overcoming} enhances continual learning through adaptive Batch Normalization, improving stability and performance. InfLoRA \cite{liang2024inflora} enhances continual learning by injecting a small number of parameters to reparameterize pre-trained weights and designing a subspace to eliminate interference between tasks. Replay-based methods include empirical replays \cite{bang2021rainbow,li2024continual,sun2022exploring, liang2023i3dod} and generative replays \cite{li2022continual,xiang2019incremental, sun2024create}. SDDR\cite{jodelet2023class} improves class-incremental learning by generating synthetic samples with Stable Diffusion for distillation and replay. Some works focus on the improvement of robots by continual learning \cite{ayub2023cbcl,gao2021cril,hafez2023continual,ayub2022few,ayub2023continual}. \cite{jiang2025continual} proposes an end-to-end multi-object grasping policy for a musculoskeletal arm with object-preference experience replay to mitigate forgetting, while LEGION \cite{meng2025preserving} introduces a Bayesian nonparametric knowledge-space framework augmented with language embeddings to accumulate and reuse lifelong RL skills for long-horizon robotic tasks.
 LOTUS \cite{wan2023lotus} stores a few human demos of novel tasks into the growing skill library, enabling lifelong learning ability for robots. M2Distill \cite{roy2025m2distill} is a multimodal distillation framework for lifelong imitation learning that enforces cross-step consistency of vision–language–action latent representations and constrains consecutive Gaussian Mixture Model policies.
However, these methods predominantly rely on 2D visual observations, and thus lack explicit 3D representations, which fundamentally limits spatial reasoning under occlusion changes and hinders robust language–action grounding for scalable language-conditioned behavior cloning. Moreover, many existing methods treat skill-wise knowledge as category-wise knowledge, overlooking its intrinsic decomposition into skill-shared and skill-specific components, which limits transfer across skills and exacerbates forgetting.

\section{Methodology}
\subsection{Problem Definition and Overview} 

\textbf{Problem Definition.} By following traditional continual learning methods \cite{rebuffi2017icarl,douillard2020podnet}, we define a multi-modal manipulation skill data stream as $\mc{T} = \{\mc{T}^m\}_{m=1}^M$, where $M$ denotes the number of continual skill learning tasks. The objective of NBAgent is to execute all acquired skills after observing $\mc{T}$. Specifically, each skill learning task consists of various robotic manipulation skills. The $m$-th continual skill learning task $\mc{T}^m = \{\mc{D}^m_i\}_{i=1}^{N^d}$ consists of $N^d$ skill demonstrations. Each skill demonstration $\mc{D}^m_i$ can be extracted to a set of keyframe actions \cite{shridhar2023perceiver}, \emph{i.e.}, $\mc{D}^m_i = \{\mf{k}^{m,i}_j\}_{j=1}^{N^k},\mf{k}^{m,i}_j=\{\mf{a}^{m,i}_{j},\mf{r}^{m,i}_{j}, \mf{l}^{m,i}\}$, where $N^k$ is the total keyframe action quantity. Given the current state in action space $\mf{a}$, the structured observation $\mf{r}$ and language instruction $\mf{l}$, the agent is expected to predict the next best keyframe action, which can be served as an action classification task \cite{james2022coarse}.  Additionally, the structured observation $\mf{r}$ is composed of the RGB-D images captured by a single front camera. An action state $\mf{a}$ can be divided into a discretized translation $\mf{a}_{tran} \in \mb{R}^3$, rotation $\mf{a}_{rot} \in \mb{R}^{(360/5)\times 3}$, gripper open state $\mf{a}_{grip} \in \{0,1\}$, and collision avoidance $\mf{a}_{col} \in \{0,1\}$. Here the rotation parameter $\mf{a}_{rot}$ entails the discretization of each rotation axis into a set of R = 5 bins. The collision avoidance parameter $\mf{a}_{col}$ provides guidance to the agent regarding the imperative need to avoid collisions.


\textbf{Overview.} The overview of NBAgent to learn skill-wise knowledge is shown in Fig.~\ref{fig: overview}. When observing a novel sequential skill learning task $\mc{T}^m$, we initialize the perceiver model $\Theta^m_p$ for the current task utilizing the model $\Theta^{m-1}_p$ obtained from the last task, and store $\Theta^{m-1}_p$ as a teacher model to compute the SRD loss. Given a RGB-D image and language instruction, as shown in Fig.~\ref{fig: overview}, $\Theta^m_p$ firstly encodes RGB-D input to obtain a deep 3D voxel via utilizing a scaled-down voxel encoder. We here design a SSR module to transfer skill-shared semantic of 3D scenes across novel and past encountered skills. Afterwards, the patched voxel and language embeddings are input to cross-attention blocks and self-attention blocks to perform feature extraction and semantics fusion, where we propose SEP to learn skill-specific knowledge from latent and low-rank space. Finally, we utilize a Q-function head $\mf{Q}^m$ to predict the state of the next keyframe in voxel space, where a SRD loss $\mc{L}_{\mathrm{SRD}}$ is developed to tackle catastrophic forgetting by aligning skill-shared voxel representation.

\subsection{Skill-Shared Semantic Rendering Module}

For language-conditional behaviour-cloning manipulation, a comprehensive semantics understanding of the 3D scenes \cite{driess2022reinforcement} plays a key role in enabling agent to perform complicated  manipulation skills. Especially in NBRL problem, which exists skill-shared semantic  across various skills, such as 3D objects and scene semantics. The existence of overlooking on common semantic space makes these semantics incomplete, further resulting in catastrophic forgetting on past learned skills. Considering this motivation, we develop a skill-shared semantic rendering module (SSR) to transfer skill-shared semantic of 3D scenes, where a NeRF model and a vision foundation model are introduced to provide semantics supervision for this transfer process.

Concretely, we draw inspiration from 3D visual representation learning \cite{ze2023gnfactor}, and leverage a latent-conditioned NeRF architecture \cite{yu2021pixelnerf}. This architecture could synthesize RGB color $\mf{c}$ of a novel image views like traditional NeRF \cite{mildenhall2021nerf}, and render the semantics $\mf{s}$ from the 3D voxel space as follows:
\begin{align}
	\label{eq: nerf}
	\mc{F}_{\Theta^m_n}(\mf{x}, \mf{d}, \mf{v}_s) =(\sigma, \mf{c}, \mf{s}),
\end{align}
where $\mc{F}_{\Theta^m_n}$ denotes the neural rendering function of NeRF model $\Theta^m_n$ in the $m$-th continual skill learning task. The 3D voxel feature $\mf{v}_s$ is obtained by a grid sample method based on trilinear interpolation from the 3D voxel observation $\mf{v}$. $\mf{x}$ and $\sigma$ are the 3D input point and differential density, and $\mf{d}$ represents unit viewing direction. The camera ray $\mf{r}$ can be obtained by: $\mf{r} = \mf{o} +t\mf{d}$, where $\mf{o}$ indicates the camera origin. By adding field-wise branches, $\Theta^m_n$ performs the same neural rendering function $\mc{F}_{\Theta^m_n}$ to estimate RGB color $\mf{c}$ and semantic feature $\mf{s}$. Therefore, the same accumulated transmittance $T(t)$ is shared to predict the two different fields and is defined as $	T(t)=\mr{exp} (- \int^t_{t_n} \sigma(s)ds)$.
In light of this, a RGB image $\mf{C}$ and 2D semantic map $\mf{M}$ can be rendered as :
\begin{align}
	\label{eq: nerf_color}
	\mf{C}(\mf{r}, \mf{v}_s)=\int^{t_f}_{t_n}T(t)\sigma(\mf{r},\mf{v}_s)\mf{c}(\mf{r},\mf{d}, \mf{v}_s)dt, \\ \mf{M}(\mf{r}, \mf{v}_s)=\int^{t_f}_{t_n}T(t)\sigma(\mf{r},\mf{v}_s)\mf{s}(\mf{r},\mf{d}, \mf{v}_s)dt.
\end{align}

\begin{algorithm}[t]			
\LinesNumbered    	\caption{Optimization Pipeline of NBAgent.} 
	\label{alg: optimization}

\textbf{Initialize:} 	Continual skill learning tasks $\{\mc{T}^m\}_{m=1}^M$ with datasets $\mc{T}^m = \{\mc{D}^m\}_{i=1}^{N^d}$; perceiver model: $\Theta^0_p$; NeRF model: $\Theta^0_n$; Pre-trained diffusion model: $\Theta_u$; Pre-trained CLIP  language encoder: $\mc{E}_c$; Memory buff: $\mc{M}=\emptyset$; Iterations: $\{\mc{I}^m\}_{m=1}^M$.  \\

\textcolor{blue}{$\triangleright$ 
\textbf{While observing a new task $\mc{T}^m$}:} \\ 
  \For {$z=1, 2, \cdots, \mc{I}^m$} 
  {Randomly select keyframe $\mf{k}^m_j$  from $\{\mathcal{D}_i^m\}_{i=1}^{N^d}  \cup \mc{M}$; \\
 
 Obtain $\mr{v}^m, \mr{v}^{m-1}, \mr{l}_s, \mr{l}_x$ utilizing $\mc{E}^m_v$, $\mc{E}^{m-1}_v$and $\mc{E}_c$;\\

 Compute $\mc{L}_{\mr{SSR}}$ by SSR ($\mr{v}^m, \mr{v}^{m-1}, \mr{l}_x,  \Theta_u $) in Eq.~\eqref{eq: ssr}; \\
 $\mf{S}[h,:],  \mf{W}_r[h,:] \xleftarrow{}$ SEP ($\mf{l}_s$); \\

Compute $\mc{L}_{\mr{CE}}, \mc{L}_{\mr{SRD}}$ utilizing $\mf{S}[h,:]$,  $\mf{W}_r[h,:]$, $\Theta^m_p,\Theta^{m-1}_p$; \\

Update $\Theta^m_p$ by Eq.~\eqref{eq: total_loss}; }
         
Store few samples from $\{\mathcal{D}_i^m\}_{i=1}^{N^d}$ in $\mc{M}$;\\
\textbf{Return}: $\Theta^m_p, \mc{M}$.

\end{algorithm}

To distill skill-shared semantic in the $m$-th skill learning task, we initialize a NeRF model acquired from the last task and denote it as $\Theta_n^{m-1}$. Then, we feed the same input to obtain the pseudo ground truth $\hat{\mf{C}}$ by Eq.~\eqref{eq: nerf_color}. We design a loss function to supervise the reconstruction process as follows:
\begin{align}
	\label{eq: color_loss}
	\nonumber \mc{L}_{\mathrm{C}} &= \underset{\mf{r}\in \mc{R}}{\sum}\Vert\mf{C}(\mf{r},\mf{v}_s)-\mf{Y}_c(\mf{r})\Vert^2_2  \\ 
 &+ \alpha \underset{\mf{r}\in \mc{R}}{\sum}\Vert\mf{C}(\mf{r},\mf{v}_s)-\hat{\mf{C}}(\mf{r},\mf{v}_s)\Vert^2_2 \cdot \mb{I}_{\mf{v}_s \notin \mc{T}^m},
\end{align}
where $\mf{Y}_c$ indicates the ground truth color and $\mc{R}$ is the set of all camera rays. $\alpha$ is the hyper-parameter to control the weight of loss function. $\mb{I}_{\mf{v}_s \notin \mc{T}^m}$ is defined such that $\mb{I}_{\mf{v}_s \notin \mc{T}^m}=1$, when the condition $\mf{v}_s \notin \mc{T}^m$ is satisfied, and $\mb{I}_{\mf{v}_s \notin \mc{T}^m}=0$ otherwise.

Considering the insufficiency in capturing skill-shared semantic by reconstructing novel views, we introduce a pre-trained visual foundation model that contains robust scene semantics to provide supervision. Relying on being pre-trained on large-scale vision-language dataset, Stable Diffusion model \cite{rombach2022high} can possess robust intrinsic representational capabilities, which is consequently utilized for semantic representation in segmentation and classification tasks \cite{xu2023open,li2023diffusion}. In light of this, we employ Stable Diffusion model $\Theta_u$ to extract vision-language semantics for supervision. Given a input view, \emph{i.e.}, $\mf{Y}_c$, we perform a one-step noise adding process to obtain a noisy image $\mf{Y}_{c,t}$. Then we utilize diffusion model $\Theta_u$ to collect vision-language semantic feature as the ground truth $\hat{\mf{F}}_s$:
\begin{align}
	\label{eq: noise_image}
	\mf{Y}_{c,t}(\mf{r}) := \sqrt{\alpha_t} \mc{E}_v(\mf{Y}_{c}(\mf{r}) )+\sqrt{1-\alpha_t}\epsilon,
\end{align}
\begin{align}
	\label{eq: semantic_label}
 \hat{\mf{F}}_s(\mf{r} ,\mf{l}_p) = \Theta_u (\mf{Y}_{c,t}(\mf{r}) , \mc{E}_c(\mf{l}_p)),
\end{align}
where $\mc{E}_v$ is a VAE encoder to encode image $\mf{Y}_c$ from pixel space to latent semantic space. $t$ represents the diffusion process step, $\epsilon \sim \mc{N}(0,1)$  and $\alpha_t$ is designed to control the noise schedule. $\mf{l}_p$ denotes the language prompt modified from task description $\mf{l}$.
To perform skill-shared semantic transfer, we align the rendered semantic feature $\mf{F}_s$ and diffusion feature $\hat{\mf{F}}_s$, and obtain the objective of our SSR module as follows:
\begin{align}
	\label{eq: ssr}
 	\mc{L}_{\mathrm{SSR}} = \mc{L}_{\mr{C}} + \lambda_{1} \mc{L}_{\mr{S}},
\end{align}
\begin{align}
	\label{eq: semantic_loss}
 \mc{L}_{\mr{S}} = \underset{\mf{r}\in \mc{R}}{\sum}\Vert\mf{F}_s(\mf{r}, \mf{v}_s)-\hat{\mf{F}}_s(\mf{r},\mf{l}_p)\Vert^2_2,
\end{align}
where $\lambda_1$ is the hyper-parameter.


\subsection{Skill-Shared Representation Distillation Module}
To address semantic overlooking on past learned skills, we develop a skill-shared representation distillation module (SRD) to align skill-shared semantic in 3D voxel representation space, as presented in Fig.~\ref{fig: overview}. Specifically, given a multi-modal keyframe input $\mf{k}^m_j=\{\mf{a}^m_{j},\mf{r}^m_{j}, \mf{l}^m\}$ from a mini-batch, we feed it into our perceiver model $\mr{\Theta}^m_p$ to obtain a keyframe on voxel space, denoted as voxel representation $\mf{v}^m_{r,j}$. The next keyframe prediction can then be computed utilizing a Q-function head as
$\mf{Q}^m (\mf{v}^m_{r,j}) =\{\mf{P}^m_{j, tran}, \mf{P}^m_{j, rot}, \mf{P}^m_{j, grip}, \mf{P}^m_{j, col} \}$, where $\mf{P}^m_{j, tran}, \mf{P}^m_{j, rot}, \mf{P}^m_{j, grip}, \mf{P}^m_{j, col}$ denote prediction on discretized translation, rotation, gripper open state and  collision avoidance. To supervise NBAgent to learn skill-wise knowledge, we follow \cite{shridhar2023perceiver}, and introduce a cross-entropy loss as :
\begin{align}
	\label{eq: cross_entropy}
	\mc
{L}_{\mr{CE}} = - \frac{1}{B}\sum^B_{j=1} \mf{Y}^m_j \mr{log} (\mf{Q}^m(\mf{v}^m_{r,j} ), ~~ \mf{v}^m_{r,j} = \Theta_p^m(k^m_j), 
\end{align}
where $B$ represents the batch size, and $\mf{Y}^m_j =\{\mf{Y}^m_{j, tran}, \mf{Y}^m_{j, rot} , \mf{Y}^m_{j, grip} , \mf{Y}^m_{j, col} \}$ is the ground truth for predicting the next keyframe. In light of this, NBAgent can continually learn skill-wise knowledge from current dataset $\mc{T}^m$ and memory buff $\mc{M}$. However, due to the limited amount of data available from old skills in memory buff $\mc{M}$, the skill-shared semantic drift on 3D voxel representation occurs between novel and old skills. It further results in forgetting skill-shared knowledge on old skills.

To address the aforementioned problems, we take an attempt to employ knowledge distillation to align voxel representation between old and new model in NBRL problem. Specifically, we initialize a teacher model with the perceiver model $\Theta^{m-1}_p$  from the last task to extract the soft label: $ \hat{\mf{Y}}^m_j = \mf{Q}^{m-1} (\mf{v}^{m-1}_{r,j}), 
    \hat{\mf{Y}}^m_j=\{\hat{\mf{Y}}^m_{j, tran}, \hat{\mf{Y}}^m_{j, rot} , \hat{\mf{Y}}^m_{j, grip} , \hat{\mf{Y}}^m_{j, col} \}$ and apply the Kullback-Leibler divergence to align the outputs of two agents as follows:
\begin{align}
	\label{eq: kl_loss}
	\mc{L}_{\mr{SRD}} = \frac{1}{\hat{B}}\sum^B_{j=1} 
\rho(\hat{\mf{Y}}^m_j/\tau)\mr{log} (\frac{\rho(\hat{\mf{Y}}^m_j/\tau)}{\rho(\mf{Q}^m(\mf{v}^m_{r,j}/\tau)})\cdot \mb{I}_{\mf{k}^m_j \notin \mc{T}^m}, 
\end{align}
where $\hat{B} = \sum_{j=1}^B\mb{I}_{\mf{k}^m_j \notin \mc{T}^m}$, $\rho$ indices the softmax function. $\tau$ is a hyper-parameter represents distillation temperature.

\subsection{Skill-Specific Evolving Planner} 
To learn category-wise knowledge over the above 3D scene representation, existing continual learning methods \cite{rebuffi2017icarl, douillard2020podnet} assume that the knowledge acquired from novel tasks and that from previous tasks are mutually independent. Differently, we consider decoupling the knowledge as the skill-shared knowledge and skill-specific knowledge, when learning skill-wise knowledge. For instance, an agent aims to ``\emph{stack the wine bottle}'' after learning how to ``\emph{open the wine bottle}''. It does not need to relearn the skill-shared knowledge of 3D scene understanding and object recognition; instead, the agent is expected to focus on acquiring the skill-specific knowledge related to the operation sequence of stacking and reducing forgetting on skill-shared knowledge. Considering this motivation, we design a skill-specific evolving planner (SEP) to perform knowledge decoupling, enabling effectively continual learning of novel manipulation skills. Specifically, inspired by \cite{wang2022dualprompt}, we first develop an adaptive language semantic bank to retrieve the skill-specific language semantic embeddings. As for these skill-specific embeddings, SEP can effectively guide the perceiver model $\Theta^{m}_p$ to encode multi-modal input from skill-specific latent and low-rank space. It results in learning the novel knowledge through the above skill-shared backbone and a light skill-specific subnetwork.

\begin{figure*}[t]
\centering
\includegraphics[width = 0.98\linewidth]
{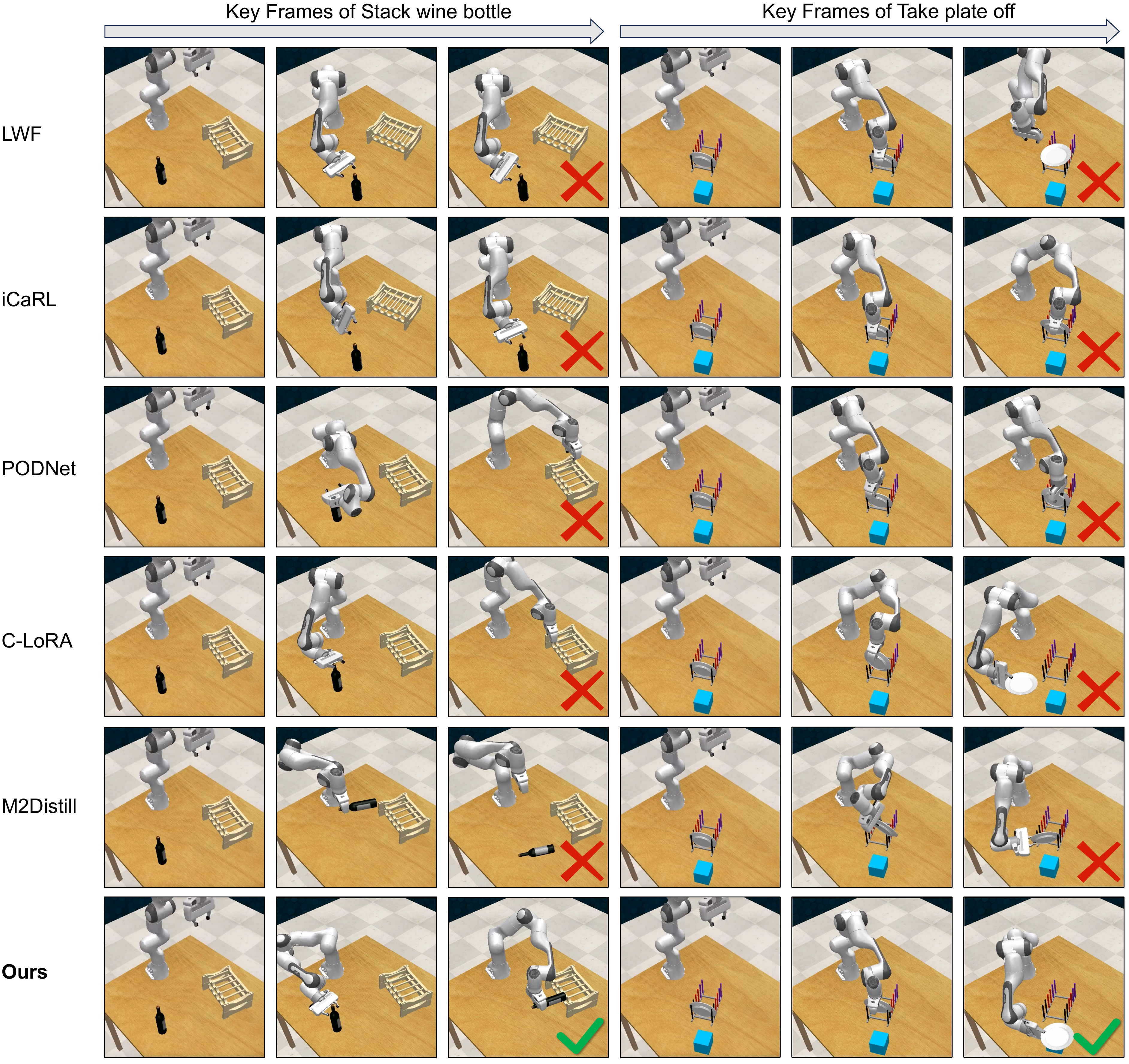}
\vspace{-6pt}
\caption{Prediction examples on RLBench \cite{james2020rlbench}. For qualitative evaluation, we visualize three key frames from each manipulation task across all methods.}
\vspace{-2pt}
\label{fig: sim_visual}
\end{figure*}

When observing a novel skill, we utilize a language encoder $\mc{E}_c$ from CLIP \cite{radford2021learning} to encode the input language instruction $\mf{l}$. Then a skill-specific language semantic embedding $\mf{l}_s \in \mb{R}^{D^s}$  can be obtained as:  $\mf{l}_s  = \mc{E}_c(\mf{l})$, where $D^s$ represents the dimension of $\mf{l}_s$. To build an adaptive language semantic bank $\mc{B}$,  we then compensate the semantic embedding $\mf{l}_s$ via an exponential moving average strategy:
\begin{align}		
\mc{B}[h,:]=(1-\mc{C}_{max})\mc{B}[h,:]+\mc{C}_{max} \mf{l}_s,
\label{eq: eq1}
\end{align} 									
where $\mc{B} \in \mb{R}^{N^b \times D^s}$ is initialized by $N^b$ zero vectors. $\mc{C} \in \mb{R}^{N^b}$ is the cosine similarity matrix between the sentence information $\mf{l}_s$ and each vector in $\mc{B}$.  if $\mc{C}_{max} > \delta$, we set $h=\mr{arg max}(\mc{C})$; otherwise, we set $h = \mr{nonzero}(\mc{B})+1$ and $\mc{C}_{max} = 1$, where $\delta=0.8$, and $\mr{nonzero}  (\cdot)$ is an operation employed to calculate the number of nonzero vectors in $\mc{B}$. In light of this, each skill corresponds to a skill-wise code $h$.

\begin{algorithm}[t]	
    	\caption{Pipeline of Our SEP.} 
	\label{alg: esp}
    
\textbf{Initialize}: Adaptive language semantic bank $\mc{B}$ with $N^b$ zero vectors; Dynamic skill-specific latent space $\mf{S}=\emptyset$; Low-rank space $\mf{W}_r=\emptyset$;  Hyper-parameter $\delta$; \\

\textcolor{blue}{$\triangleright$ 
\textbf{Input}:} language embedding $\mf{l}_s$; \\
 Compute cosine similarity matrix $\mc{C}$ between $\mc{B}$ and $\mf{l}_s$; \\
       
\If{$\mc{C}_{max}>\delta$} {$h \xleftarrow{} \mr{argmax}(\mc{C});$ \\

Update $\mc{B}$ by Eq.~(\ref{eq: eq1}); \\   

\textbf{Return}: $\mf{S}[h,:],  \mf{W}_r[h,:]$;} 

\Else{
 $h \xleftarrow{} \mr{nonzero}(\mc{B})+1;$ \\
 Expand $\mc{B}$ by Eq.~(\ref{eq: eq1});  \\     
 Randomly initialize $\mf{S}[h,:], \mf{W}_r[h,:]$; \\
 
 \textbf{Return}: $\mf{S}[h,:],  \mf{W}_r[h,:]$;}

\textcolor{blue}{$\triangleright$ 
\textbf{Output}:}  $\mf{S}[h,:],  \mf{W}_r[h,:]$. \\
\end{algorithm}

To better learn skill-specific knowledge, we utilize the skill-wise code $h$ to guide the network to perform skill-specific network training. Although existing multi-modal behaviour-cloning methods \cite{shridhar2023perceiver,ze2023gnfactor}  encode multi-modal input from a skill-shared latent space, we here consider developing a dynamic skill-specific latent space $\mf{S} \in \mb{R}^{N^s\times N^l \times D}$ to facilitate skill-specific knowledge learning, where $N^s$,  $N^l$,  $D$ denote the number of learned skills, the learnable latent vector quantity and the vector dimension.  By following \cite{shridhar2023perceiver}, we encode these latent vectors in $\mf{S}$ with the multi-modal input to obtain the cross-attention features $\mf{F}_c \in \mb{R}^{(N^p+N^e)\times D}$ as follows:
\begin{align}
	\label{eq: cross_attention}
	\!\! \mathbf{F}_c =  \rho(\frac{\mr{Cat}(\mathbf{p}, \mf{e})\mathbf{W}_q (\mathbf{S}[h,:] \mathbf{W}_k)^{\top}}{\sqrt{d}})(\mathbf{S}[h,:] \mathbf{W}_v), 
\end{align}
where $\mathbf{W}_q, \mathbf{W}_k, \mathbf{W}_v\in \mb{R}^{D\times D}$ are linear projection layers, and $\mr{Cat}(\cdot)$ denotes the concatenation operation. $d$ is a scaling factor. $\mf{p} \in \mb{R}^{N^p \times D}$ is obtained by concatenating the current state of agent and the input voxelized observation. $\mf{e} \in \mb{R}^{N^e \times D}$ is the encoded embeddings of the input language instruction. To this end, relying on the skill-wise code $h$, NBAgent can continually encode the novel multi-modal input from a skill-specific latent space, which is beneficial to learn skill-specific knowledge from latent space.

\begin{figure*}[t]

\includegraphics[width=360pt]
{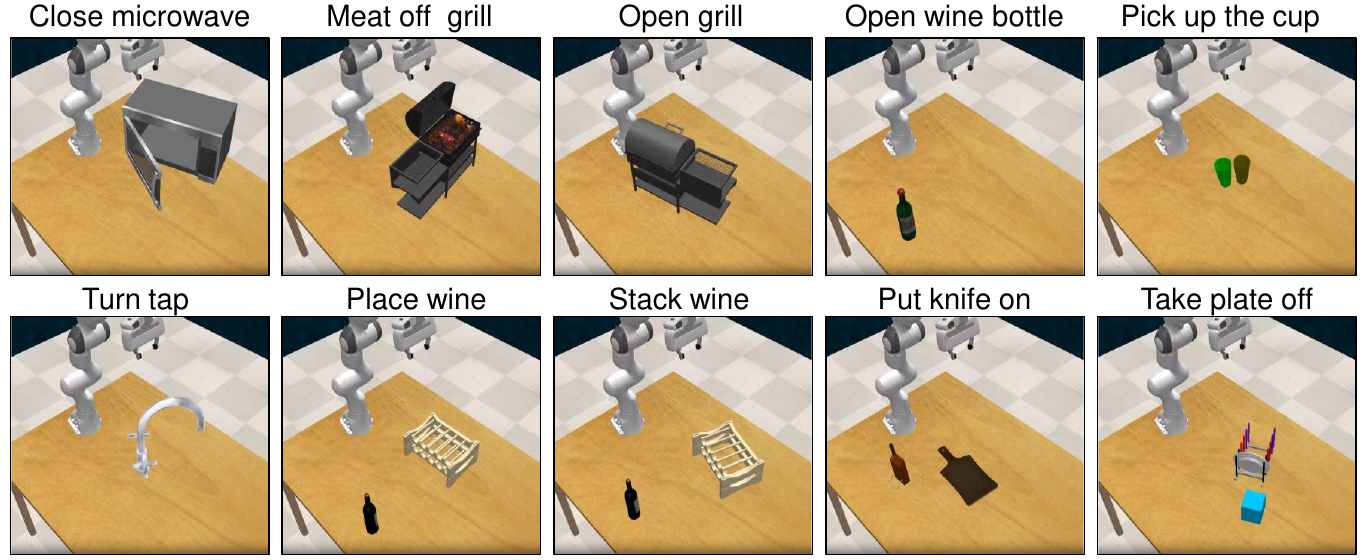}
\centering
\vspace{-6pt}
\caption{Visualization of manipulation skills in dataset Kitchen, consisting of 10 manipulation
skills pertinent to kitchen environments.} 
\label{fig: kitchen}
\vspace{-2pt}
\end{figure*}

\begin{figure*}[!t]
\centering
\includegraphics[width=380pt]
{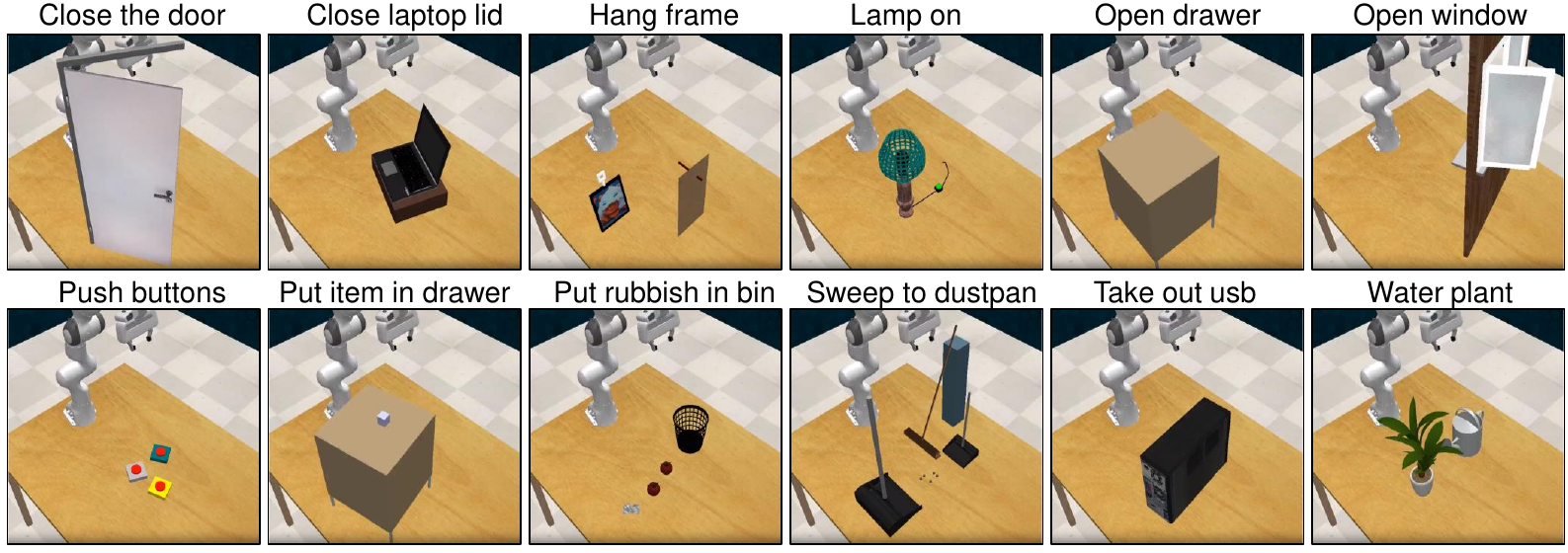}
\vspace{-6pt}
\caption{Visualization of manipulation skills in dataset Living Room, including 12 manipulation
skills associated with living
room scenarios. } 
\label{fig: living_room}
\vspace{-2pt}
\end{figure*}

Considering the limitation in representing skill-specific knowledge from latent space, we further explore to learn skill-specific knowledge from low-rank space.  Specifically, we introduce a low-rank adaptation layer (LoRA) \cite{hu2021lora} that can learn skill-specific  knowledge in an efficient manner. For $\mathbf{W}_q, \mathbf{W}_k, \mathbf{W}_v, \mathbf{W}_o$ in each attention block, we design a set of skill-specific LoRA layers to perform skill-specific forward and obtain the final output feature $\mf{F}_x$ as follows:
\begin{align}
	\label{eq: lora}
	\mathbf{F}_x =  \mf{X} \mf{W}+ \mf{X} \mf{W}_{r} [h,:] = \mf{X} \mf{W} + \mf{X} \mf{W}_a[h,:] \mf{W}_b[h,:],
\end{align}
where $\mf{W} \in \mb{R}^{D\times D}$ represents one of $\mathbf{W}_q, \mathbf{W}_k, \mathbf{W}_v, \mathbf{W}_o$ and $\mf{X} \in \mb{R}^{(N^p+N^e)\times D}$ denotes the input feature for these projection layers. $\mf{W}_r[h,:] = \mf{W}_a [h,:]\mf{W}_b [h,:]$ is a low-rank decomposition, where $\mf{W}_a \in \mb{R}^{ D \times N^r}$ is initialized in a  random Gaussian manner,  $\mf{W}_b \in \mb{R}^{N^r\times D }$ is initialized by zero and $N^r$ is a hyper-parameter controlling the size of LoRA layers. On the one hand, obviously, $\mf{W}$ is shared among all skills and expected to learn skill-shared knowledge.  On the other hand, each  $\mf{W}_r$ is executed to learn the corresponding skills and perform skill-specific forward with the guidance of skill-wise code $h$, resulting in continually embedding skill-specific knowledge to our NBAgent from low-rank space. Specifically, we summary the process of our SEP in \textbf{Algorithm} \ref{alg: esp}.

In conclusion, both SSR and SRD modules above are designed to capture the 3D geometric and semantic information related to the skill, and further transfer skill-shared semantic amongst diverse manipulation skills to tackle past skill forgetting. To perform skill-specific knowledge learning, we develop the SEP to accumulate skill-specific knowledge in latent and low-rank space, thereby effectively learning novel skills. Finally, the objective function of our NBAgent can be simplified as:
\begin{align}
	\label{eq: total_loss}
	\mc{L}_{\mr{Total}} = \mc{L}_{\mr{CE}} + \mc{L}_{\mr{SSR}}+ \lambda_{2} \mc{L}_{\mr{SRD}},
\end{align}
where $\lambda_{2}$ represents hyper-parameter. $\Theta^m_p, \Theta^m_n$ are optimized via $\mathcal{L}_{\mr{CE}}$ and $\mathcal{L}_{\mr{SSR}}$  for the first task, and trained via $\mathcal{L}_{\mr{Total}}$ in Eq.~(\ref{eq: total_loss}) when $m\geq2$.

\begin{table}[t]
\centering
\begin{center}
\vspace{-4pt}
\setlength{\tabcolsep}{0.3mm}
\setlength{\arraycolsep}{1.0pt}
\renewcommand{\arraystretch}{1.15}
\caption{\textbf{Skill Details in Kitchen \& Living Room}}
\vspace{-15pt}
\label{tab: skill_details}
\resizebox{1.0\linewidth}{!}{
\begin{tabular}{cllcc}
\hline
    \textbf{ID} &\textbf{Skill in Kitchen } &\textbf{Language Example} &\textbf{VA.}  &\textbf{Avg. Kf}   \\
\hline
1&close microwave &``close microwave" &1 &2.3 \\
2&meat off grill &``take the steak off the grill" &2 &6.0 \\	
3&open grill &``open the grill" &1 &4.5	\\
4&open wine bottle &``open wine bottle" &1 &3.3	 \\
5&pick up cup &``pick up the red cup" &20 &3.8 \\
6&turn tap &``turn left tap" &2 &3.0 \\
7&place wine &``stack the wine bottle" &3 &6.1 \\
8&put knife on &``put the knife on the board" &1 &5.1 \\
9&stack wine &``stack wine bottle" &1 &7.4 \\
10&take plate off &``take plate off colored rack" &3 &6.3 \\
\bottomrule
 \textbf{ID} &\textbf{Skill in Living Room } &\textbf{Language Example} &\textbf{VA.} &\textbf{Avg. Kf} \\
\hline
1&close door &``close the door" &1 &4.8 \\
2&close laptop lid &``close laptop lid" &1 &6.0 \\
3&hang frame &``hang frame on hanger" &1 &5.3 \\
4&lamp on &``turn on the light" &1 &3.5 \\
5&open drawer &``open the bottom drawer" &3 &4.8 \\
6&open window &``open left window"  &1 &6.0 \\
7&push buttons &``push the white button"  &18 &4.8 \\
8&put item in drawer &``put the item in the top drawer" &3 &14.3 \\
9&put rubbish in bin &``put rubbish in bin" &1 &5.0 \\
10&sweep to dustpan &``sweep dirt to the tall dustpan" &2 &6.2 \\
11&take usb out &``take usb out of computer" &1 &3.3  \\
12&water plants &``pour some water on the plant" &1 &6.3  \\
\bottomrule
\end{tabular}}
\end{center}
\vspace{-10pt}
\end{table}



\section{Experiments}
\label{experiments}

We conduct experiments to investigate how NBAgent leverages both skill-shared and skill-specific knowledge learning to address the NBRL problem on two simulation datasets. To this end, we design five experimental paradigms: (1) comparison performance on RLBench using simulation data to validate overall learning capability (Sec.~\ref{sec: comparison} and Sec.~\ref{sec: skill-wise}),  (2) ablation studies on key modules and hyperparameters to quantify their individual contributions (Sec.~\ref{sec: ablation}), (3) task complexity analysis to examine performance under varying skill complexities (Sec.~\ref{sec: task_complexity}), and (4) computational complexity analysis to assess training and inference efficiency (Sec.~\ref{sec: computation_complexity}).


\subsection{Implementation Details}
\label{implementation_datails}

\textbf{Dataset.} By following PerAct \cite{shridhar2023perceiver}, we conduct our experiments on RLBench \cite{james2020rlbench} and simulate robot learning in CoppelaSim \cite{rohmer2013v}. Here, language annotations are obtained through manually designed templates, while key frames are extracted by monitoring the robot’s state and task-related events during simulation. To simulate the working scenarios of NBRL, we design two NBRL benchmark datasets,  called \textcolor{deepred}{\textbf{Kitchen}} and \textcolor{deepred}{\textbf{Living Room}}. Specifically, as shown in Figs.~\ref{fig: kitchen}-\ref{fig: living_room} and Tab.~\ref{tab: skill_details}, Kitchen dataset is constructed by gathering 10 manipulation skills pertinent to kitchen environments, and Living Room dataset consists of 12 manipulation skills associated with living room scenarios. Each manipulation skill includes a training set of 20 episodes and a test set of 25 episodes. Furthermore, these skills involve various variations encompassing randomly sampled attributes such as colors, sizes, counts, placements, and object categories, resulting in a total of 69 distinct variations. In this section, we employ a fixed skill identification scheme, where skills are uniquely indexed within the ranges of 1-10 and 1-12 for distinct operational scenarios.  A comprehensive summary of these skills, including their functional descriptions, associated key frame statistics and variation quality, is presented in Tab.~\ref{tab: skill_details}. Specifically, the \textbf{ID} column denotes the unique identifier for each skill, while the \textbf{Avg. Kf} metric quantifies the mean number of key frames observed across 20 training episodes. Additionally, \textbf{Va.} denoting the number of variations per skill, provides a summary of the dataset and reflects the diversity and complexity of the contained scenarios.



\begin{table*}[t]
\centering
\setlength{\tabcolsep}{0.6mm}
\setlength{\arraycolsep}{2.5pt}
\renewcommand{\arraystretch}{1.1}
\normalsize
\caption{Comparisons of success rate (\%) on Kitchen and Living Room. \textcolor{deepred}{\textbf{Red}} and \textcolor{blue}{\textbf{Blue}} represents the highest results and runner-up. The number of steps  equals the sum of quantity of base skill learning task and continual skill learning tasks.}
\scalebox{0.93}{
\begin{tabular}{l|ccccc|ccccc|ccccc|ccccc}
	\toprule
\makecell[c]{\multirow{2}{*}{Comparison Methods}}  & \multicolumn{5}{c|}{\textbf{5-5}  $(2~\mr{steps})$} & \multicolumn{5}{c|}{\textbf{5-1} $(6~\mr{steps})$} & \multicolumn{5}{c|}{\textbf{6-3}  
 $(3~\mr{steps})$} & \multicolumn{5}{c}{\textbf{6-2} $(4~\mr{steps})$}\\
	  &1-5 &6-10 &All &Avg. &For. &1-5 &6-10 &All &Avg. &For. &1-6 &7-12 &All &Avg. &For. &1-6 &7-12 &All &Avg. &For.\\
	\midrule
    PerAct \cite{shridhar2023perceiver} (CoRL'2023) &58.9 &30.1 &44.5 &\textbf{$\mathrm{-}$} &\textbf{$\mathrm{-}$} &58.9 &30.1 &44.5 &\textbf{$\mathrm{-}$} &\textbf{$\mathrm{-}$} &34.7 &27.3 &31.0 &\textbf{$\mathrm{-}$} &\textbf{$\mathrm{-}$} &34.7 &27.3 &31.0 &\textbf{$\mathrm{-}$} &\textbf{$\mathrm{-}$}\\
    GNFactor \cite{ze2023gnfactor} (CoRL'2023)  &56.3 &32.3 &44.3 &\textbf{$\mathrm{-}$} &\textbf{$\mathrm{-}$} &56.3 &32.3 &44.3 &\textbf{$\mathrm{-}$} &\textbf{$\mathrm{-}$}  &48.0 &32.0 &40.0 &\textbf{$\mathrm{-}$} &\textbf{$\mathrm{-}$} &48.0 &32.0 &40.0 &\textbf{$\mathrm{-}$} &\textbf{$\mathrm{-}$}\\
 \midrule
    Fine-Tuning &15.7 &26.4 &21.1 &38.9 &41.1 &3.2 &20.0 &9.6 &13.8 &39.1  &4.4 & 29.8 & 17.1 &29.0 &44.1 &6.2 &19.1 &12.7 &24.2 &43.9    \\
    ER \cite{chaudhry2018riemannian} (ECCV'2018) &\textcolor{blue}{\textbf{56.0}} &25.6 &40.8 &50.0 &\textcolor{blue}{\textbf{3.2}} &53.6 &29.6 &41.6 &49.7 &9.8 &43.8 &31.1 &37.4 &42.2 &\textcolor{deepred}{\textbf{7.7}} &40.7 &30.2 &35.4 &38.1 &17.6 \\
    LWF \cite{li2017learning} (TPAMI'2017) &\textcolor{deepred}{\textbf{57.6}} &30.4 &\textcolor{blue}{\textbf{44.0}} &\textcolor{blue}{\textbf{51.0}} &\textcolor{deepred}{\textbf{1.8}} &52.0 &30.2 &41.1 &50.4 &11.5 &40.0 &32.7 &36.3 &42.9 &13.5 &43.2 &31.4 &37.2 &41.0 &15.8 \\
    iCaRL \cite{rebuffi2017icarl} (CVPR'2017) &42.4 &\textcolor{blue}{\textbf{34.4}} &38.4 &49.8 &15.2 &\textcolor{blue}{\textbf{54.0}} &33.4 &\textcolor{blue}{\textbf{43.7}} &51.0 &9.8 &42.0 &32.0 &37.0 &43.1 &11.8 &44.0 &30.8 &37.4 &41.5 &15.4 \\
    PODNet \cite{douillard2020podnet} (ECCV'2020) &52.8 &30.4 &41.6 &50.6 &10.8 &53.6 &31.2 &42.4 &\textcolor{blue}{\textbf{53.2}} &\textcolor{blue}{\textbf{9.6}} &43.4 &32.7 &38.1 &43.9 &11.6 &\textcolor{blue}{\textbf{44.7}} &32.0 &\textcolor{blue}{\textbf{38.4}} &\textcolor{blue}{\textbf{42.2}} &\textcolor{blue}{\textbf{12.7}} \\
    C-LoRA \cite{smith2024continualdiffusion} (TMLR'2024) &48.0 &32.8 &40.4 &40.2 &13.6 &49.6 &\textcolor{blue}{\textbf{35.2}} &42.4 &52.5 &10.3  &42.0 &\textcolor{blue}{\textbf{39.3}} &40.7 &44.6 &\textcolor{blue}{\textbf{9.6}} &40.7 &\textcolor{blue}{\textbf{34.7}} &37.7 &41.8 &14.7\\
    M2Distill \cite{roy2025m2distill} (ICRA'2025)   &55.5 
 &30.9 	&43.2 	&\textcolor{blue}{\textbf{51.0}} 	&8.3  &53.1 	&34.4 	&\textcolor{blue}{\textbf{43.7}} 	&52.8 	&10.7  &\textcolor{deepred}{\textbf{45.8}} 	&36.5 	&\textcolor{blue}{\textbf{41.1}} 	&\textcolor{blue}{\textbf{45.8}} 	&10.2 &43.6 	&28.0 	&\textcolor{blue}{\textbf{38.4}} 	&40.8 	&15.1

  \\
        \midrule
        \rowcolor{lightgray}
    \textbf{Ours} &53.6 &\textcolor{deepred}{\textbf{36.3}} &\textcolor{deepred}{\textbf{44.9}} &\textcolor{deepred}{\textbf{52.5}} &6.4 &\textcolor{deepred}{\textbf{54.4}}   &\textcolor{deepred}{\textbf{37.6}} &\textcolor{deepred}{\textbf{46.0}} &\textcolor{deepred}{\textbf{55.2}} &\textcolor{blue}{\textbf{8.9}} &\textcolor{blue}{\textbf{44.9}} &\textcolor{deepred}{\textbf{42.2}} &\textcolor{deepred}{\textbf{43.6}} &\textcolor{deepred}{\textbf{47.6}} &\textcolor{deepred}{\textbf{7.7}} &\textcolor{deepred}{\textbf{45.8}} &\textcolor{deepred}{\textbf{35.3}} &\textcolor{deepred}{\textbf{40.6}} &\textcolor{deepred}{\textbf{43.5}} &\textcolor{deepred}{\textbf{10.9}}\\
        \midrule
\end{tabular}}
\label{tab: Comparison}
\vspace{-5pt}
\end{table*}

\begin{figure}[t]
\centering
\includegraphics[width=1.0\linewidth]
{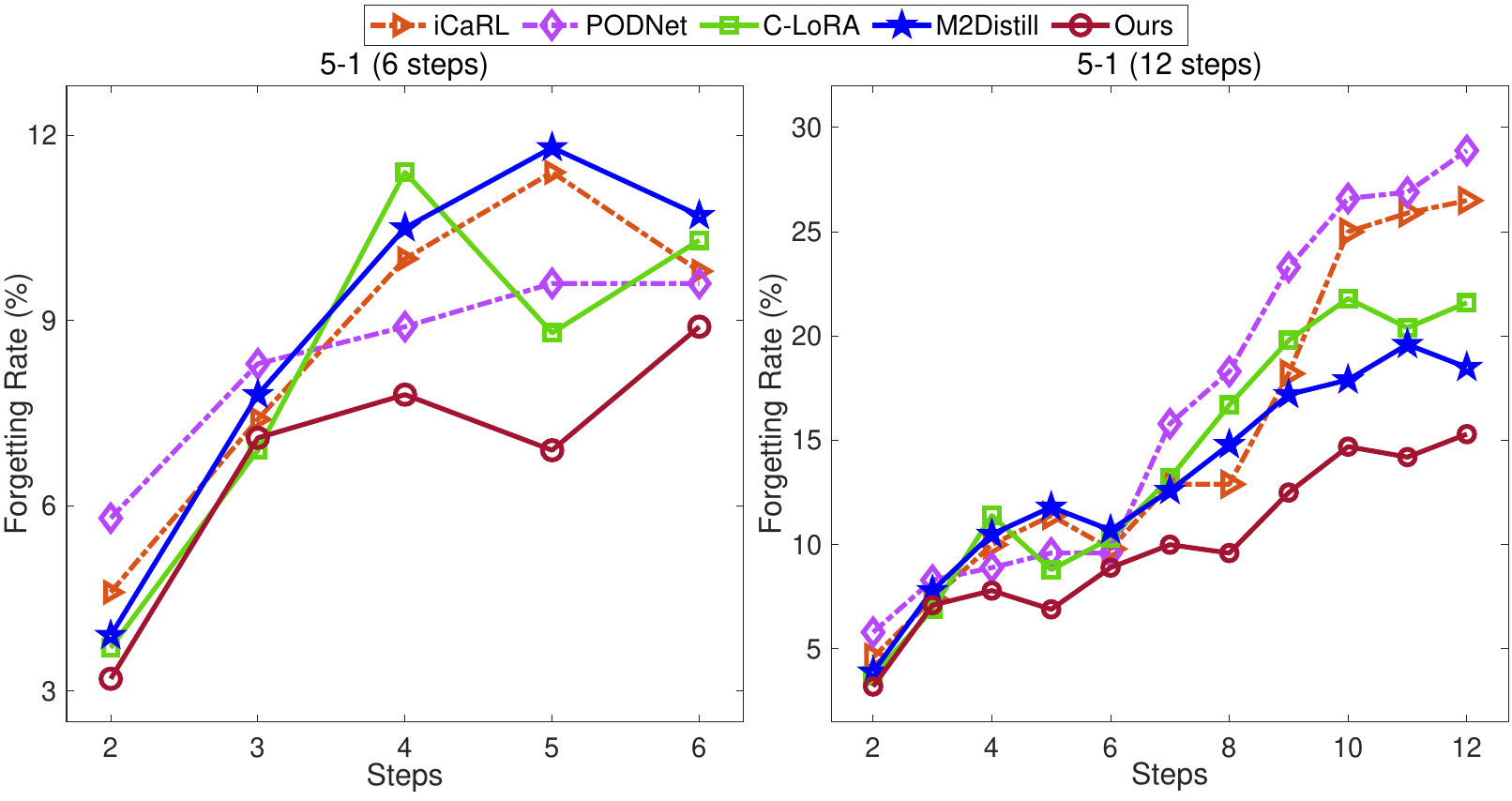}
\vspace{-23pt}
\caption{Comparisons of forgetting rate  (\emph{i.e.}, For.) (\%) under the settings of 5+1 (6 steps) and 5+1 (12 steps).} 
\label{fig: forgetting_rate}
\vspace{-4pt}
\end{figure}

\textbf{Baselines.} We conduct the comprehensive evaluation between our NBAgent and the following two multi-task behavior-cloning methods: PerAct \cite{shridhar2023perceiver}, GNFactor \cite{ze2023gnfactor}, and five continual learning methods in behavior-cloning agent setting: Fine-Tuning, ER \cite{chaudhry2019tiny}, LWF \cite{li2017learning}, iCaRL \cite{rebuffi2017icarl}, PODNet \cite{douillard2020podnet} and C-LoRA \cite{smith2024continualdiffusion}. More specifically, both PerAct \cite{shridhar2023perceiver} and GNFactor\cite{ze2023gnfactor} joint train all manipulation skills within one-stage training, for two datasets respectively, referred to as the upperbound. Fine-Tuning enables continual learning by updating all parameters of the Perceiver model for novel skills. ER \cite{chaudhry2018riemannian} randomly stores old skills to memory buff and replay them when learning novel skills. LWF \cite{li2017learning} applies knowledge distillation between current and old models using replayed samples. iCaRL \cite{rebuffi2017icarl} introduces memory management to select and update stored episodes in the memory buffer. C-LoRA \cite{smith2024continualdiffusion} employs an orthogonal regularization loss to prevent catastrophic forgetting during task-specific LoRA adaptation. M2distill \cite{roy2025m2distill}  preserves a consistent latent space across vision, language, and action modalities to mitigate catastrophic forgetting.

\begin{table*}[t]
\centering
\setlength{\tabcolsep}{2.15mm}
\setlength{\arraycolsep}{1.1pt}
\normalsize
\caption{Comparison results  on Kitchen under the setting of \textbf{5-5}.}
\scalebox{0.92}{
\begin{tabular}{l|cccccccccc|ccc|c}
	\toprule
\makecell[c]{\multirow{1}{*}{Skill ID}} 
	  &1 &2 &3 &4 &5 &6 &7 &8 &9 &10 &1-5 &6-10 &All. &Imp.  \\
	\midrule
    PerAct \cite{shridhar2023perceiver} (CoRL'2023) &96.0 &77.3 &53.3 &30.7 &37.3 &52.0 &2.7 &18.7 &4.0 &73.3 &58.9 &30.1 &44.5 &$\Uparrow$ 0.4\\
    GNFactor \cite{ze2023gnfactor} (CoRL'2023) &92.0 &60.0 &70.7 &12.0 &46.7 &52.0 &5.3 &22.7 &10.7 &70.7 &56.3 &32.3 &44.3 &$\Uparrow$ 0.6\\
 \midrule
    Fine-Tuning   &10.7 &0.0 &33.3 &33.3 &1.3 &49.3 &6.7 &14.7 &5.3 &56.0 &15.7 &26.4 &21.1 &$\Uparrow$ 23.8\\
    ER \cite{chaudhry2018riemannian} (ECCV'2018) &96.0 &72.0 &56.0 &18.7 &37.3 &45.3 &0.0 &14.7 &1.3 &66.7 &\textcolor{blue}{\textbf{56.0}} &25.6 &40.8 &$\Uparrow$ 4.1 \\
     LWF \cite{li2017learning} (TPAMI'2017) &97.3 
 &72.0 &54.7 &33.3 &33.3 &48.0 &6.7 &34.7 &4.0 &56.0 &\textcolor{deepred}{\textbf{57.6}} &30.4  &42.8 &$\Uparrow$ 2.1  \\
    iCaRL \cite{rebuffi2017icarl} (CVPR'2017) &60.0 &41.3  &77.3 &22.7 &32.0 &61.3 &2.7 &21.3 &10.7 &81.3 &42.4 &\textcolor{blue}{\textbf{34.4}} &38.4 &$\Uparrow$ 6.5     \\ 
    PODNet \cite{douillard2020podnet} (ECCV'2020) &93.3 &74.7 &42.7 &24.0 &28.0 &41.3 &9.3 &32.0 &6.7 &64.0 &52.8 &30.4 &41.6 &$\Uparrow$ 3.3\\
    C-LoRA \cite{smith2024continualdiffusion} (TMLR'2024) &74.7 &40.0 &56.0 &33.3 &36.0 &49.3 &17.3 &24.0 &12.0 &61.3 &48.0 &32.8 &40.4 &$\Uparrow$ 4.5 \\
    M2Distill \cite{roy2025m2distill} (ICRA'2025) &84.0 	&70.7 	&56.0 	&29.3 	&37.3 	&52.0 	&0.0 	&29.3 	&9.3 	&64.0 	&55.5 	&30.9 	&\textcolor{blue}{\textbf{43.2}} &$\Uparrow$ 1.7
 \\
        \midrule
        \rowcolor{lightgray}
        \textbf{Ours} &94.7 &52.0 &38.7 &44.0 &38.7 &64.0 &2.7 &34.7 &9.3 &70.7 &53.6 &\textcolor{deepred}{\textbf{36.3}} &\textcolor{deepred}{\textbf{44.9}} &\textbf{$\mathrm{-}$} \\
        \midrule
\end{tabular}}
\label{tab: kitchen-5-5}
\vspace{-5pt}
\end{table*}

\begin{table*}[t]
\centering
\setlength{\tabcolsep}{2.15mm}
\setlength{\arraycolsep}{1.1pt}
\normalsize
\caption{ Comparison results on Kitchen  under the setting of \textbf{5-1}.}
\scalebox{0.92}{
\begin{tabular}{l|cccccccccc|ccc|c}
	\toprule
\makecell[c]{\multirow{1}{*}{Skill ID}} 
	  &1 &2 &3 &4 &5 &6 &7 &8 &9 &10 &1-5 &6-10 &All &Imp.  \\
	\midrule
    PerAct \cite{shridhar2023perceiver} (CoRL'2023) &96.0 &77.3 &53.3 &30.7 &37.3 &52.0 &2.7 &18.7 &4.0 &73.3 &58.9 &30.1 &44.5 &$\Uparrow$ 1.5\\
    GNFactor \cite{ze2023gnfactor} (CoRL'2023) &92.0 &60.0 &70.7 &12.0 &46.7 &52.0 &5.3 &22.7 &10.7 &70.7 &56.3 &32.3 &44.3 &$\Uparrow$ 1.7\\
 \midrule
    Fine-Tuning   &17.3 &0.0 &0.0 &0.0 &0.0 &8.0  &0.0 &0.0 &9.3 &64.0 &3.2 &20.0 &9.7  &$\Uparrow$ 36.3\\
    ER \cite{chaudhry2018riemannian} (ECCV'2018) &90.7	&54.7 	&37.3	&54.7	&28.0 	&60.0	 &5.3	&0.0	&24.0	&62.7 &53.6 &29.6 &41.6 &$\Uparrow$ 4.4
 \\
     LWF \cite{li2017learning} (TPAMI'2017) &88.0 &50.7 &42.7 &41.3 &37.3 &53.3 &10.7 &14.7 &9.3 &64.0 &52.0 &30.2 &41.1  &$\Uparrow$ 4.9\\
    iCaRL \cite{rebuffi2017icarl} (CVPR'2017) &89.3 &54.7 &56.0 &32.0 &38.7 &49.3 &8.0 &29.3 &14.7 &68.0 &\textcolor{blue}{\textbf{54.0}} &33.4 &\textcolor{blue}{\textbf{43.7}}  &$\Uparrow$ 2.3\\
    PODNet \cite{douillard2020podnet} (ECCV'2020) &92.0  &57.3 &53.3 &32.0 &34.7 &52.0 &14.7 &24.0 &13.3 &50.7 &53.6 &31.2 &42.4  &$\Uparrow$ 3.6\\
    C-LoRA \cite{smith2024continualdiffusion} (TMLR'2024) &80.0 &49.3 &50.7 &29.3 &40.0 &69.3 &18.7 &16.0 &14.7 &57.3 &49.6 &\textcolor{blue}{\textbf{35.2}} &42.4 &$\Uparrow$ 3.6\\

    M2Distill \cite{roy2025m2distill} (ICRA'2025) &88.0 	&56.0 	&49.3 	&32.0 	&40.0 	&70.7 	&14.7 	&17.3 	&14.7 	&54.7 & 53.1 	&34.4 	&\textcolor{blue}{\textbf{43.7}} &$\Uparrow$ 2.3

 \\

        \midrule
        \rowcolor{lightgray}
        \textbf{Ours}  &92.0 &61.3 &44.0  &34.7	&41.3 &76.0	&17.3	&26.7	&14.7	&52.0 &\textcolor{deepred}{\textbf{54.4}} &\textcolor{deepred}{\textbf{37.6}} &\textcolor{deepred}{\textbf{46.0}} &\textbf{$\mathrm{-}$}
    \\
        \midrule

\end{tabular}}
\label{tab: kitchen5-1}
\vspace{-5pt}
\end{table*}

\textbf{Training Details.} 
In NBRL, following continual learning setting \cite{douillard2020podnet}, we assume that the agent undergoes initial learning through a set of manipulation skills, referred to as the base skill learning task. Subsequently, skills are added incrementally in steps, characterized as continual skill learning tasks. To valuate the robustness of NBAgent in  addressing NBRL problem, we set up four combinations of base skills and new skills: 5-5, 5-1, 6-3, and 6-2, which are highlighted in bold in Tab.~\ref{tab: Comparison}. For instance, ``5-1'' indicates that the agent is initially trained on a base skill learning task comprising the first five manipulation skills in the skill ID. Subsequently, the agent undergoes five continual skill learning tasks, each involving a new skill, ultimately learning a total of 10 skills in the Kitchen dataset. 
In NBRL-R dataset, we adopt an NBRL setting denoted as ``4-1'', where the agent first learns the initial four skills listed in Tab~\ref{tab: simpler_details} and then continually learns one new skill over 3 steps. Additionally, we use RT-1 \cite{brohan2022rt} as the base behavior-cloning agent  and apply the proposed NBAgent, along with other comparison methods, to perform NBRL. To evaluate the long-term learning capability of NBAgent against other methods, we design a novel long-term learning setting, denoted as 5-1 (12 steps). In this setting, the agent is initially trained on the first five manipulation skills from the Kitchen dataset and subsequently undergoes 11 continual learning tasks, each introducing a new skill drawn from the remaining five Kitchen skills and the first six Living Room skills.

\begin{figure}[t]
\centering
\includegraphics[width=1.0\linewidth]
{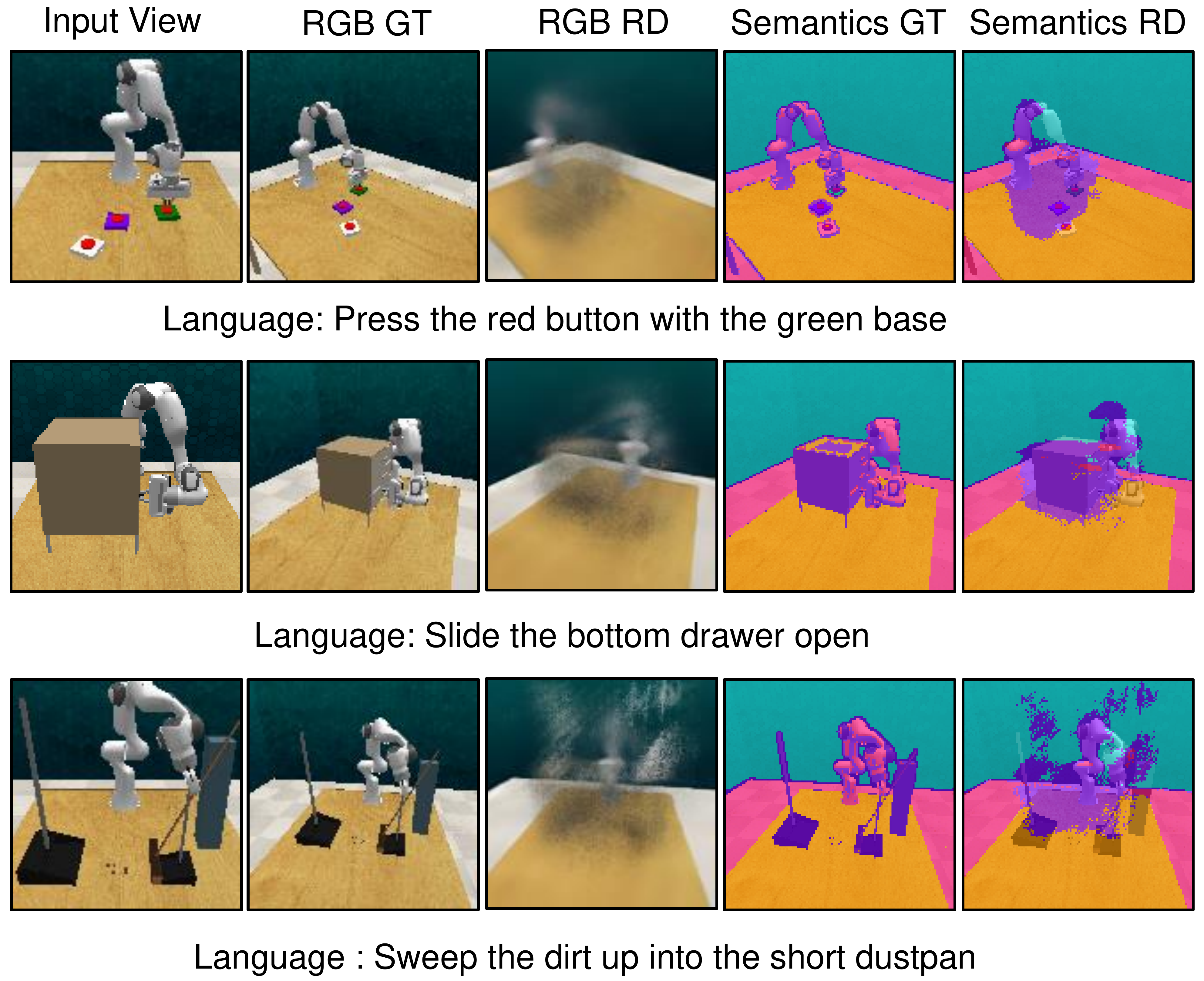}
\vspace{-20pt}
\caption{Visualization of rendering results in our SSR module. RGB GT denotes the color ground truth, and Semantics GT represents the semantics ground truth extracted by Stable Diffusion model. RGB RD and Semantics RD are the rendering novel view and semantic feature.} 

\label{fig: render_visual}
\vspace{-5pt}
\end{figure}

 In addition, the LAMB \cite{you2019large} optimizer is applied for all methods with a initial learning rate of $5.0 \times 10^{-4}$ and a batch size of 2. We utilize 100K training iterations for PerAct and GNFactor, 80K for base task training, 20K for incremental task training. We  store a fixed 4 episodes of each old skill in $\mc{M}$ for all methods. We configure the hyper-parameters as $\alpha=0.1$ in Eq.~(\ref{eq: color_loss}), $\delta = 0.8$ in SEP module, with $\lambda_1 = 0.1$ and
$\lambda_2 = 0.2$ to balance $\mc{L}_S$ in Eq.~(\ref{eq: ssr}) and $\mc{L}_{SRD}$ in Eq.~(\ref{eq: total_loss}).

\textbf{Evaluation Metric.} Following behaviour-cloning methods \cite{ze2023gnfactor, goyal2023rvt}, we adopt the success score (\%) $\mc{I}^m_{i}$ as the basic indicator for evaluation, where $\mc{I}^m_{i}$ represents the success score of $i$-th skill tested at $m$-th continual skill learning task.  Specifically, after learning the final skill learning task,  we compute the mean success score of the skills learned at the base skill learning task (Base) $\mc{I}^M_{B}$, continual skill learning task (Novel) $\mc{I}^M_{N}$ and all skill learning tasks (All) $\mc{I}^M_{A}$. These metrics respectively reflect the robustness of old skill forgetting , the capacity of novel skill learning, as well as its overall performance. Additionally, we introduce a Avg. metric $G$ and For. metric $F$ to measure average performance and skill-wise forgetting rate over the whole NBRL process, where $G = \frac{1}{M} \sum_{m=1}^M \mc{I}^m_A$ and $ F=  \frac{1}{N^m} \sum^{N^m}_{i=1} \underset{m \in \{1, ..., M-1\}}{\mr{max}}( \mc{I}^m_{i} - \mc{I}^M_{i})$.

\begin{table*}[t]
\centering
\setlength{\tabcolsep}{1.47mm}
\setlength{\arraycolsep}{1.1pt}
\vspace{-4pt}
\normalsize
\caption{Comparison results  on Living Room  under the setting of \textbf{6-3}.}
\scalebox{0.92}{
\begin{tabular}{l|cccccccccccc|ccc|c}
	\toprule
\makecell[c]{\multirow{1}{*}{Skill ID}} 
	  &1 &2 &3 &4 &5 &6 &7 &8 &9 &10 &11 &12 &1-6 &7-12 &All &Imp.  \\
	\midrule
    PerAct \cite{shridhar2023perceiver} (CoRL'2023)  &5.3 &66.7 &0.0 &84.0 &38.7 &13.3 &16.0 &2.7 &41.3 &0.0 &97.3 &6.7 &34.7 &27.3 &31.0 &$\Uparrow$ 12.6 \\
    GNFactor \cite{ze2023gnfactor} (CoRL'2023)  &2.7 &92.0 &1.3	&84.0 &80.0 &28.0 &8.0 &1.3	&46.7 &32.0	&100 &0.0 &48.0 &32.0 &40.0 &$\Uparrow$ 3.6\\
 \midrule
    Fine-Tuning  &8.0 &18.7	&0.0 &0.0 &0.0 &0.0 &0.0 &0.0 &0.0 &68.0 &100.0 &10.7 &4.4 &29.8 &17.1 &$\Uparrow$ 25.6\\
    ER \cite{chaudhry2018riemannian} (ECCV'2018) &10.7 &58.7 &24.0 &81.3 &76.0 &12.0 &5.3 &4.0 &78.7 &8.0 &89.3 &1.3 &\textcolor{blue}{\textbf{43.8}} &31.1 &37.4 &$\Uparrow$ 6.2 \\
     LWF \cite{li2017learning} (TPAMI'2017) &16.0 &62.7 &9.3 &82.7 &52.0 &17.3  &4.0 &9.3 &52.0 &33.3 &84.0 &14.7 &40.0 &32.7 &36.3  &$\Uparrow$ 7.3\\
    iCaRL \cite{rebuffi2017icarl} (CVPR'2017) &29.3 &80.0 &21.3 &74.7 &36.0 &10.7 &4.0 &2.7 &68.0 &1.3 &96.0 &20.0 &42.0 &32.0  &37.0 &$\Uparrow$ 6.6 \\ 
    PODNet \cite{douillard2020podnet} (ECCV'2020) &13.3 &56.0 &26.7 &78.7 &61.3 &25.3 &1.3 &6.7 &72.0 &8.0 &82.7 &22.7 &43.4 &32.7  &38.1   &$\Uparrow$ 5.5\\
    C-LoRA \cite{smith2024continualdiffusion} (TMLR'2024) &16.0 &53.3 &25.3 &74.7 &52.0 &32.0 &6.7 &25.3 &72.0 &14.7 &82.7 &33.3 &42.0 &\textcolor{blue}{\textbf{39.3}} &40.7 &$\Uparrow$ 2.9 \\
    M2Distill \cite{roy2025m2distill} (ICRA'2025) &13.3 	&74.7 	&28.0 	&80.0 	&54.7 	&24.0 	&10.7 	&13.3 	&62.7 	&14.7 	&88.0 	&29.3 	&45.8 	&36.5 	&\textcolor{blue}{\textbf{41.1}} &$\Uparrow$ 2.5
\\
        \midrule
        \rowcolor{lightgray}
    \textbf{Ours} &17.3 &74.7 &9.3 &82.7 &62.7 &22.7 &22.7 &9.3	&85.3	&9.3 &98.7 &28.0 &\textcolor{deepred}{\textbf{44.9}} &\textcolor{deepred}{\textbf{42.2}} &\textcolor{deepred}{\textbf{43.6}} &\textbf{$\mathrm{-}$}\\
        \midrule

\end{tabular}}
\label{tab: living6-3}
\vspace{-5pt}
\end{table*}

\begin{table*}[!t]
\centering
\setlength{\tabcolsep}{1.47mm}
\setlength{\arraycolsep}{1.1pt}
\normalsize
\caption{Comparison results  on Living Room under the setting of \textbf{6-2}.}
\scalebox{0.92}{
\begin{tabular}{l|cccccccccccc|ccc|c}
	\toprule
\makecell[c]{\multirow{1}{*}{Skill ID}} 
	  &1 &2 &3 &4 &5 &6 &7 &8 &9 &10 &11 &12 &1-6 &7-12 &All &Imp.  \\
	\midrule
    PerAct \cite{shridhar2023perceiver} (CoRL'2023)  &5.3 &66.7 &0.0 &84.0 &38.7 &13.3 &16.0 &2.7 &41.3 &0.0 &97.3 &6.7 &34.7 &27.3 &31.0 &$\Uparrow$ 9.6 \\
    GNFactor \cite{ze2023gnfactor} (CoRL'2023)   &2.7 &92.0 &1.3	&84.0 &80.0 &28.0 &8.0 &1.3	&46.7 &32.0	&100 &0.0 &48.0 &32.0 &40.0 &$\Uparrow$ 0.6\\
 \midrule
    Fine-Tuning &0.0 &37.3 &0.0 &0.0 &0.0 &0.0 &0.0 &0.0 &0.0 &0.0 &96.0 &18.7 &6.2 &19.1 &12.7 &$\Uparrow$ 27.9\\
    ER \cite{chaudhry2018riemannian} (ECCV'2018) &4.0 &89.3 &17.3 &80.0 &38.7 &14.7 &20.0 &2.7 &60.0 &4.0 &90.7 &4.0 &40.7 &30.2 &35.4 &$\Uparrow$ 5.2\\

     LWF \cite{li2017learning} (TPAMI'2017) &12.0 &53.3 &45.3 &60.0 &61.3 &33.3 &4.0 &10.7 &73.3 &5.3 &84.0 &5.3 &43.2 &31.4 &37.2 &$\Uparrow$ 3.4\\
    iCaRL \cite{rebuffi2017icarl} (CVPR'2017) &17.3 &52.0 &45.3 &68.0 &64.0 &17.3 &2.7 &24.0 &70.7 &10.7 &72.0 &4.0 &44.0 &30.8 &37.4 &$\Uparrow$ 3.2\\
    PODNet \cite{douillard2020podnet} (ECCV'2020) &10.7 &54.7 &52.0 &57.3 &60.0 &32.0 &9.3 &14.7 &70.7 &14.7 &72.0 &13.3  &\textcolor{blue}{\textbf{44.7}} &32.0 &\textcolor{blue}{\textbf{38.4}} &$\Uparrow$ 2.2\\
    C-LoRA \cite{smith2024continualdiffusion} (TMLR'2024) &16.0 &48.0 &45.3 &62.7 &52.0 &18.7 &17.3 &33.3 &70.7 &14.7 &62.7 &10.7 &40.7 &\textcolor{blue}{\textbf{34.7}} &37.7 &$\Uparrow$ 2.9\\
    M2Distill \cite{roy2025m2distill} (ICRA'2025) &16.0 	&54.7 	&48.0 	&81.3 	&61.3 	&32.0 	&9.3 	&9.3 	&64.0 	&10.7 	&70.7 	&4.0 	&43.6 	&28.0 	&\textcolor{blue}{\textbf{38.4}}   &$\Uparrow$ 2.2
\\
        \midrule
        \rowcolor{lightgray}
    \textbf{Ours} &8.0 &61.3 &32.0 &84.0 &58.7 &30.7 &4.0 &17.3 &80.0 &6.7 &98.7 &5.3 &\textcolor{deepred}{\textbf{45.8}} &\textcolor{deepred}{\textbf{35.3}} &\textcolor{deepred}{\textbf{40.6}} &\textbf{$\mathrm{-}$}\\
        \midrule
\end{tabular}}
\label{tab: living6-2}
\vspace{-5pt}
\end{table*}










\begin{table}[t]
\centering
\setlength{\tabcolsep}{1.mm}
\caption{Ablation studies on the Kitchen dataset under the settings of \textbf{5-5} and \textbf{5-1}.} 
\resizebox{0.95\linewidth}{!}{
\begin{tabular}{c|c|ccccc}
\toprule
\multirow{2}{*}{}   & \multirow{2}{*}{Variants} & \multirow{2}{*}{Base} & \multirow{2}{*}{Base+SSR} & Base+SSR & Base+SSR & ~~\multirow{2}{*}{\textbf{Ours}}~~ \\
& & & & \qquad+SEP & \qquad+SRD &  \\
\midrule
 \multirow{4}{*}{Modules} 
& SSR & \xmarkg & \cmark & \cmark & \cmark & \cmark  \\

 & SEP & \xmarkg & \xmarkg & \cmark & \xmarkg & \cmark  \\
& SRD & \xmarkg & \xmarkg & \xmarkg & \cmark & \cmark  \\
\midrule
\multirow{5}{*}{ \makecell[c]{\textbf{5-5} \\   $(2~\mr{steps})$}}
& 1-5 & 48.6 & \textcolor{deepred}{\textbf{56.0}} & 41.1 & \textcolor{blue}{\textbf{54.5}} & 53.6
 \\
& 6-10  & 26.4 &26.7  & \textcolor{deepred}{\textbf{38.1}} &28.6  & \textcolor{blue}{\textbf{36.3}} \\
&ALL  &37.5 &41.3 &39.6  &\textcolor{blue}{\textbf{41.6}} &\textcolor{deepred}{\textbf{44.9}} \\
&Avg. &45.8 &49.1 &49.8 &\textcolor{blue}{\textbf{51.2}} &\textcolor{deepred}{\textbf{52.5}} \\
&For. &18.4 &0.8 &18.9 &\textcolor{deepred}{\textbf{5.8}} &\textcolor{blue}{\textbf{6.4}}\\
\midrule

\multirow{5}{*}{\makecell[c]{\textbf{5-1} \\   $(6~\mr{steps})$ }}
& 1-5 & 52.0 &\textcolor{blue}{\textbf{56.8}} &48.0 &\textcolor{deepred}{\textbf{57.2}} &54.4
 \\
& 6-10  & 27.8 & 30.4 &\textcolor{deepred}{\textbf{41.6}} &30.8 &\textcolor{blue}{\textbf{37.6}} \\
&ALL &40.3 &43.6 &\textcolor{blue}{\textbf{44.8}} &44.0  &\textcolor{deepred}{\textbf{46.0}} \\
&Avg. &49.4 &50.2 &\textcolor{blue}{\textbf{53.9}} &53.1 &\textcolor{deepred}{\textbf{55.2}} \\
&For. &10.6  &\textcolor{blue}{\textbf{7.1}}  &14.7 &\textcolor{deepred}{\textbf{6.5}} &8.9 \\

       \midrule

\end{tabular}
}
\label{tab: ablation_kitchen}
\vspace{-5pt}
\end{table}

\begin{table}[t]
\centering
\setlength{\tabcolsep}{1.mm}
\caption{Ablation studies on the Living Room dataset under the settings of \textbf{6-3} and \textbf{6-2}.} 
\resizebox{0.95\linewidth}{!}{
\begin{tabular}{c|c|ccccc}
\toprule
\multirow{2}{*}{}   & \multirow{2}{*}{Variants} & \multirow{2}{*}{Base} & \multirow{2}{*}{Base+SSR} & Base+SSR & Base+SSR & ~~\multirow{2}{*}{\textbf{Ours}}~~ \\
& & & & \qquad+SEP & \qquad+SRD &  \\
\midrule
 \multirow{4}{*}{Modules} 
& SSR & \xmarkg & \cmark & \cmark & \cmark & \cmark  \\

 & SEP & \xmarkg & \xmarkg & \cmark & \xmarkg & \cmark  \\
& SRD & \xmarkg & \xmarkg & \xmarkg & \cmark & \cmark  \\
\midrule
\multirow{5}{*}{ \makecell[c]{\textbf{6-3} \\   $(3~\mr{steps})$}}
& 1-6 &41.8 &42.0 &41.1 &\textcolor{deepred}{\textbf{46.6}} &\textcolor{blue}{\textbf{44.9}}
 \\
& 7-12  &31.6 &33.1 &\textcolor{blue}{\textbf{34.4}} &32.8 &\textcolor{deepred}{\textbf{42.2}} \\
&ALL &36.7 &37.6 &37.8 &\textcolor{blue}{\textbf{39.7}} &\textcolor{deepred}{\textbf{43.6}} \\
&Avg. &42.4 &45.3 &45.9 &\textcolor{blue}{\textbf{46.4}} &\textcolor{deepred}{\textbf{47.6}} \\
&For. &11.9 &12.6 &15.6 &\textcolor{deepred}{\textbf{6.8}} &\textcolor{blue}{\textbf{7.7}} \\
\midrule

\multirow{5}{*}{\makecell[c]{\textbf{6-2} \\   $(4~\mr{steps})$ }}
& 1-6 &40.3 &38.4 &40.4 &\textcolor{blue}{\textbf{44.7}} & \textcolor{deepred}{\textbf{45.8}}
 \\
& 7-12  &30.6 &\textcolor{blue}{\textbf{35.6}} &\textcolor{deepred}{\textbf{39.1}} &33.8 & 35.3 \\
&ALL &35.5 &37.0 &\textcolor{blue}{\textbf{39.8}} &39.2 &\textcolor{deepred}{\textbf{40.6}}  \\
&Avg. &39.2 &40.9 &\textcolor{blue}{\textbf{43.3}} &42.0 &\textcolor{deepred}{\textbf{43.5}}  \\
&For. &14.1 &16.5 &19.1 &\textcolor{blue}{\textbf{11.6}} &\textcolor{deepred}{\textbf{10.9}} \\

       \midrule

\end{tabular}
}
\label{tab: ablation_living}
\vspace{-5pt}
\end{table}

\subsection{Comparison Performance on RLbench}
\label{sec: comparison}


We present comparative results between NBAgent and baseline methods on Kitchen and Living Room datasets in Tabs.~\ref{tab: Comparison}-\ref{tab: living6-2} and Fig.~\ref{fig: sim_visual}. 
As shown in Tab.~\ref{tab: Comparison}, NBAgent significantly outperforms compared methods by $1.5\% \sim 41.4\%$ in terms of Avg. metric on two Kitchen dataset settings and $1.3\% \sim 19.3\%$ on two Living Room dataset settings. Additionally, NBAgent surpasses the SOTA method M2Distill \cite{roy2025m2distill} by $1.5\%\sim2.4\%$ on the Kitchen settings and $1.8\%\sim2.7\%$ on the Living Room settings in terms of the Avg. metric.   These results indicate that NBAgent  effectively addressing catastrophic  forgetting of old skills through skill-specific and skill-shared knowledge learning. 
From the comparative analysis, we derive the following observations: 1) Superiority to Learn Novel Skills: Despite observing the skill data streams, NBAgent  surprisingly 
 surpasses  the joint training methods PerAct \cite{shridhar2023perceiver} and GNFactor \cite{ze2023gnfactor}  by $0.4\%\sim 1.7\%$ in average success rate across all skills, providing additional evidence to support the efficacy of our model. 2) Robust Forgetting Mitigation: Throughout the whole NBRL process, our NBAgent exhibits $1.9\%\sim6.7\%$ lower forgetting metrics than all comparison methods on Living Room 6-2 setting, highlighting its robustness in retaining learned skills.

In addition, we illustrate key simulation inference frames as qualitative results  in Fig.~\ref{fig: sim_visual}. We can find that  comparison methods frequently fail to complete manipulation tasks due to two critical limitations: 1) inability to accurately recognize and localize objects, and 2) suboptimal action cloning caused by catastrophic forgetting.  In contrast, NBAgent addresses these challenges better, confirming its effectiveness in preserving skill-specific and skill-shared knowledge.

\subsection{Skill-Wise Comparison Performance}
\label{sec: skill-wise}
We presented the skill-wise comparative results in Tabs.~\ref{tab: kitchen5-1}, \ref{tab: living6-3}, \ref{tab: kitchen-5-5} and \ref{tab: living6-2}.  The experimental results reveal that our model achieves the highest average skill-wise success rate, demonstrating improvements of $2.1\% \sim 36.3\%$ on two Kitchen dataset settings and $2.2\% \sim 25.6\%$ on two Living Room dataset settings, compared to other methods. 
These findings further validate the effectiveness of our framework in addressing the challenges of the NBRL problem. NBAgent achieves superior performance over baseline methods in most skill-wise comparisons, highlighting its capability to mitigate catastrophic forgetting through dual mechanisms: 1) Skill-Shared Knowledge Preservation: The proposed SSR and SRD modules facilitate effective transfer of 3D scene semantics between between old and current models. 2) Skill-Specific Knowledge Adaptation: The SEP module enables continual acquisition of novel skills while maintaining skill-specific attribute.

\begin{table}[t]
\centering
\setlength{\tabcolsep}{0.8mm}
\setlength{\arraycolsep}{1.6pt}
\renewcommand{\arraystretch}{1.3}
\normalsize
\caption{Comparison results in terms of success rate (\%) on Living Room dataset when setting the various size of memory buff $\mc{M}$.}
\scalebox{0.90}{
\begin{tabular}{c|ccccc|ccccc}
	\toprule
\makecell[c]{\multirow{2}{*}{Buffer size}}  & \multicolumn{5}{c|}{\textbf{6-3} $(2~\mr{steps})$} & \multicolumn{5}{c}{\textbf{6-2} $(4~\mr{steps})$} \\
	  &1-6 &7-12 &All &Avg. &For. &1-6 &7-12 &All &Avg. &For.\\
	\midrule
    $|\mc{M}|=2$  &42.0 &41.3 &41.7 &44.9 &\textcolor{blue}{\textbf{4.4}} &32.0 &30.7 &31.3 &37.2 &18.8 \\
    $|\mc{M}|=4$  &\textcolor{blue}{\textbf{44.9}} &\textcolor{deepred}{\textbf{42.2}} &\textcolor{deepred}{\textbf{43.6}} &\textcolor{deepred}{\textbf{47.6}} &7.7 &\textcolor{blue}{\textbf{45.8}} &\textcolor{deepred}{\textbf{35.3}} &\textcolor{blue}{\textbf{40.6}} &\textcolor{blue}{\textbf{43.5}} &\textcolor{blue}{\textbf{10.9}} \\
    $|\mc{M}|=6$  &\textcolor{deepred}{\textbf{50.0}} &\textcolor{blue}{\textbf{36.7}} &\textcolor{blue}{\textbf{43.3}} &\textcolor{blue}{\textbf{45.0}} &\textcolor{deepred}{\textbf{2.2}} &\textcolor{deepred}{\textbf{50.0}} &\textcolor{blue}{\textbf{34.7}} &\textcolor{deepred}{\textbf{42.3}} &\textcolor{deepred}{\textbf{45.4}} &\textcolor{deepred}{\textbf{8.0}} \\
        \midrule

\end{tabular}}
\label{tab: replay}
\vspace{-5pt}
\end{table}

\begin{figure}[!t]
\centering
\includegraphics[width=1.0\linewidth]
{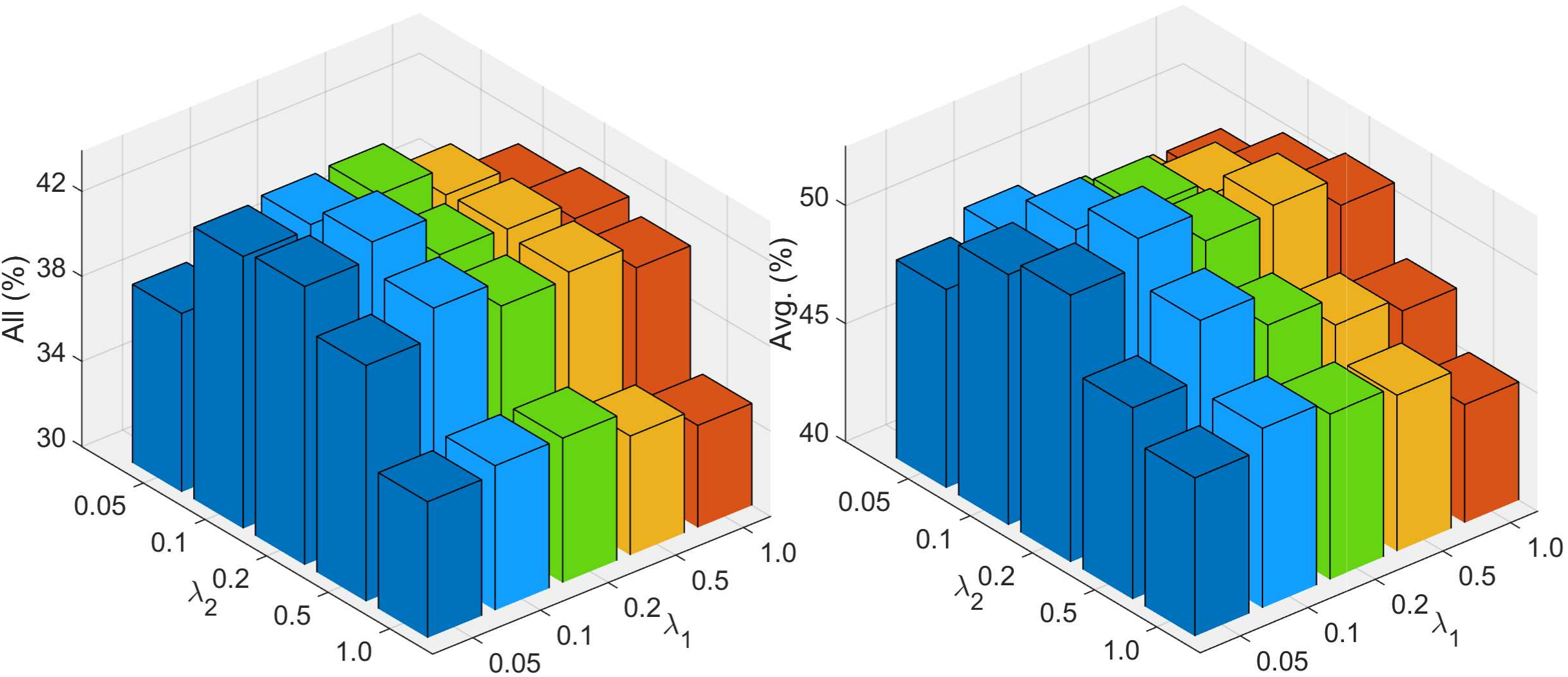}
\vspace{-20pt}
\caption{Analysis of hyper-parameters $\lambda_1$ and $\lambda_2$ under the 5-5 setting.}

\label{fig: hyperparam}
\vspace{-5pt}
\end{figure}

\begin{table}[t]
\centering
\setlength{\tabcolsep}{0.8mm}
\setlength{\arraycolsep}{1.6pt}
\renewcommand{\arraystretch}{1.3}
\normalsize
\caption{Comparison results in terms of success rate (\%) on Living Room dataset when setting the various task orders.}
\scalebox{0.90}{
\begin{tabular}{c|ccccc|ccccc}
	\toprule
\makecell[c]{\multirow{2}{*}{Task order}}  & \multicolumn{5}{c|}{\textbf{6-3} $(2~\mr{steps})$} & \multicolumn{5}{c}{\textbf{6-2} $(4~\mr{steps})$} \\
	  &1-6 &7-12 &All &Avg. &For. &1-6 &7-12 &All &Avg. &For.\\
	\midrule
    Order 1  &44.9 &42.2 &43.6 &47.6 &7.7 &45.8 &35.3 &40.6 &43.5 &10.9\\
    Order 2  &41.4 &43.5 &42.5 &46.1 &8.4 &40.6 &38.5 &40.0 &41.8 &8.6  \\
    Order 3  &45.6 &42.1 &43.9 &48.4 &7.5 &45.9 &37.9 &41.9 &44.0 &8.5 \\
        \midrule

\end{tabular}}
\label{tab: order}
\vspace{-5pt}
\end{table}



\subsection{Ablation Studies}
\label{sec: ablation}

\textbf{Ablation on skill-shared attribute}.
 Compared to Ours, the scores of Ours-w/oSRD on old skills are dropped by $3.8\%\sim12.5\%$ (\emph{i.e.}, 1-5 and 1-6 in Tab.~\ref{tab: Comparison}). This indicates that SRD can effectively learn skill-shared knowledge to tackle semantic overlooking on old  skills,
 To evaluate the effectiveness of SSR module, we visualize the rendering results in Fig.~\ref{fig: render_visual}. It suggests that SSR module can efficiently transfer skill-shared semantics of 3D scenes, thereby achieving an improvement about $0.5\%\sim3.1\%$ in terms of Avg compared Ours-w/oSEP\&SRD with ER in Tab.~\ref{tab: Comparison}.  The above mentioned demonstrates that NBAgent can significantly tackle old skill forgetting by preserving skill-shard semantics of 3D scenes and voxel representation.

\begin{figure}[!t]
\centering
\includegraphics[width=1.0\linewidth]
{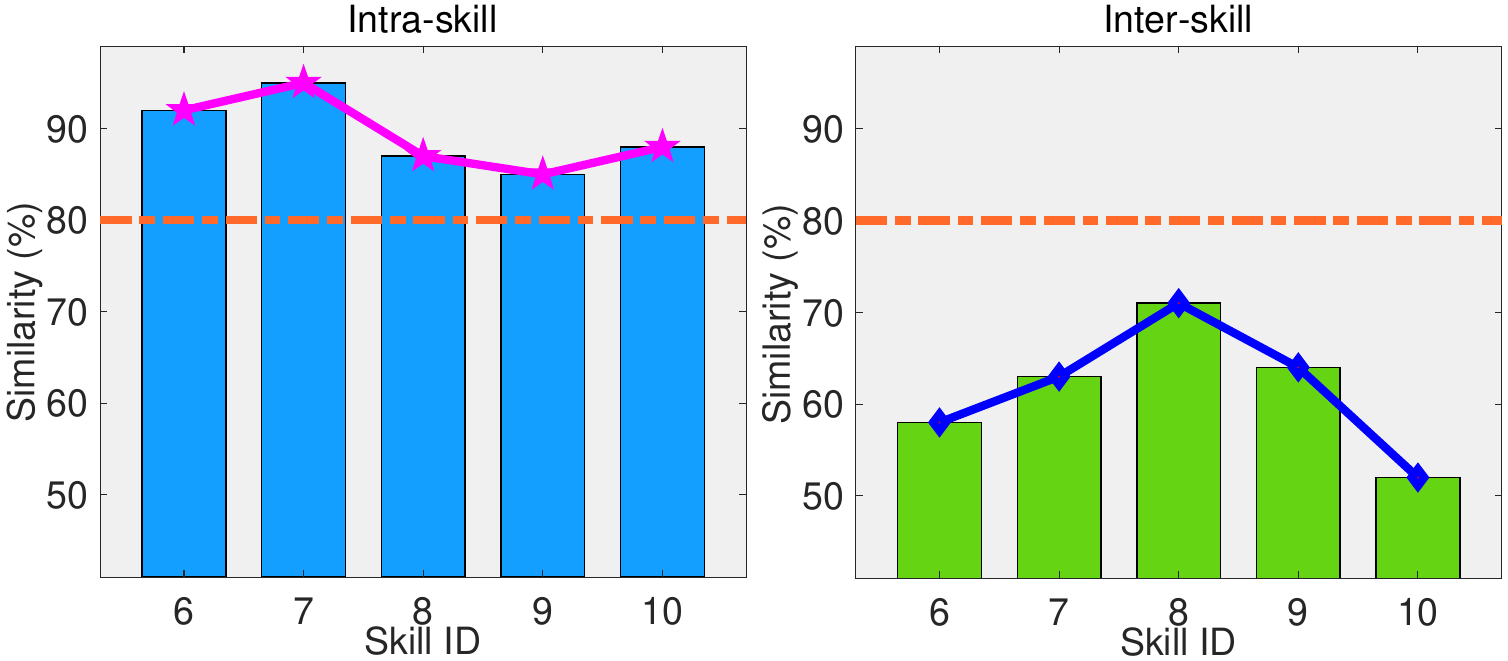}
\vspace{-22pt}
\caption{Analysis of hyper-parameters $\delta$ under the 5-5 setting.} 
\label{fig: delta}
\vspace{-5pt}
\end{figure}

\begin{table}[t]
\centering
\setlength{\tabcolsep}{1.mm}
\caption{Ablation studies on the external models in SSR module under the settings of \textbf{5-5}, where the external models including CLIP \cite{radford2021learning}, DINO \cite{caron2021emerging} and SD (Stable Diffusion) \cite{rombach2022high}. }
\resizebox{1.0\linewidth}{!}{
\begin{tabular}{c|c|cccc}
\toprule
\multirow{1}{*}{}   & \multirow{1}{*}{Variants} & \multirow{1}{*}{Base} & \multirow{1}{*}{Base+DINO} & Base+CLIP & ~~\multirow{1}{*}{\textbf{Ours}}~~ \\
\midrule
 \multirow{4}{*}{Modules} 
& DINO & \xmarkg & \cmark & \xmarkg  & \xmarkg  \\

 & CLIP  & \xmarkg & \xmarkg & \cmark  & \xmarkg  \\
& SD & \xmarkg & \xmarkg & \xmarkg  & \cmark  \\
\midrule
\multirow{5}{*}{ \makecell[c]{\textbf{5-5} \\   $(2~\mr{steps})$}}
& 1-5 & 49.4 &50.4 &\textcolor{blue}{\textbf{52.8}} &\textcolor{deepred}{\textbf{53.6}}
 \\
& 6-10  &34.8 &34.6  & \textcolor{blue}{\textbf{36.0}}  & \textcolor{deepred}{\textbf{36.3}} \\
&ALL  &42.1 &42.5   &\textcolor{blue}{\textbf{44.4}} &\textcolor{deepred}{\textbf{44.9}} \\
&Avg. &49.4  &50.2 &\textcolor{blue}{\textbf{51.6}} &\textcolor{deepred}{\textbf{52.5}} \\
&For. &10.6 &10.8 &\textcolor{blue}{\textbf{8.2}} &\textcolor{deepred}{\textbf{6.4}}\\
\midrule

\end{tabular}
}
\label{tab: ablation_ssr}
\vspace{-5pt}
\end{table}

\begin{table}[t]
\centering
\setlength{\tabcolsep}{2.6mm}
\setlength{\arraycolsep}{3.0pt}
\renewcommand{\arraystretch}{1.1}
\normalsize
\caption{Comparison results in long-term learning under the setting of \textbf{5-1} (12 steps).}
\resizebox{\linewidth}{!}{

\begin{tabular}{l|ccccc}
	\toprule
\makecell[c]{\multirow{1}{*}{Comparison Methods}} 
	  &1-5 &6-16 &All &Avg. &For.  \\
	\midrule
    Fine-Tuning  &	1.6 	&	9.1 	&	6.8 	&	9.2 	&	58.4 
  \\
   iCaRL \cite{rebuffi2017icarl} (CVPR'2017) &	36.0 	&	28.6 	&	30.9 	&	36.4 	&	26.5 
 \\
      PODNet \cite{douillard2020podnet} (ECCV'2020)  &	36.8 	&	26.4 	&	29.7 	&	35.2 	&	28.9 
   \\ 
      C-LoRA \cite{smith2024continualdiffusion} (TMLR'2024)  &\textcolor{blue}{\textbf{41.6}} 	&	30.4 	&	33.9 	&	38.7 	&	21.6 
 \\
     M2Distill \cite{roy2025m2distill} (ICRA'2025) &	40.8 	&\textcolor{blue}{\textbf{36.8}} 	&\textcolor{blue}{\textbf{38.1}} 	&\textcolor{blue}{\textbf{41.7}} 	&\textcolor{blue}{\textbf{18.5}} 
\\
        \midrule
        \rowcolor{lightgray}
    \textbf{Ours} 	&\textcolor{deepred}{\textbf{43.6}} 	&\textcolor{deepred}{\textbf{40.4}} 	&\textcolor{deepred}{\textbf{41.4}} 	&\textcolor{deepred}{\textbf{44.5}} 	&\textcolor{deepred}{\textbf{15.3}}
  \\
        \midrule

\end{tabular}}
\label{tab: longterm}
\vspace{-5pt}
\end{table}

\begin{table}[t]
\centering
\setlength{\tabcolsep}{1.6mm}
\setlength{\arraycolsep}{1.5pt}
\renewcommand{\arraystretch}{1.05}
\normalsize
\caption{Training parameters, memory cost, training time per 100 iterations and comparison results in terms of All. under the setting of \textbf{6-3}.}
	\resizebox{\linewidth}{!}{
\begin{tabular}{l|cccc}
	\toprule
\makecell[c]{\multirow{2}{*}{Comparison Methods}} 
	  &Trainable &Memory & Training &\makecell[c]{\multirow{2}{*}{All}}  \\
          &Parameters &Storage &Time (\textit{100 it}/s)  \\
	\midrule
    
    Fine-Tuning  &48.3M &0.2G &51.2s	&17.1   \\
    ER \cite{chaudhry2018riemannian} (ECCV'2018) &48.3M &5.1G &56.4s &37.4  \\
     LWF \cite{li2017learning} (TPAMI'2017) &48.3M &5.1G &78.6s &36.3 \\
    iCaRL \cite{rebuffi2017icarl} (CVPR'2017) &48.3M &5.1G  &81.4s &37.0   \\ 
    PODNet \cite{douillard2020podnet} (ECCV'2020) &48.3M &5.1G &72.7s  &38.1 \\
    C-LoRA \cite{smith2024continualdiffusion} (TMLR'2024) &18.1M &4.8G &60.6s &40.7 \\

     M2Distill \cite{roy2025m2distill} (ICRA'2025) &48.3M &4.8G &92.3s	&\textcolor{blue}{\textbf{41.1}} 
\\
        \midrule
        \rowcolor{lightgray}
    \textbf{Ours} &18.1M &4.8G &76.4s &\textcolor{deepred}{\textbf{43.6}}  \\

        \midrule

\end{tabular}}
\label{tab: parameter}
\vspace{-5pt}
\end{table}

 \textbf{Ablation on skill-specific attribute}. To evaluate effectiveness of our NBAgent on learning skill-specific knowledge, we eliminate our proposed SEP module and present results in Tabs.~\ref{tab: Comparison}, \ref{tab: kitchen5-1} and \ref{tab: living6-3}.  Ours-w/oSRD outperforms Ours-w/oSEP\&SRD on novel skills by $1.3\%\sim11.4\%$, (\emph{i.e.}, 6-10 and 7-12 in Tab.~\ref{tab: Comparison}), which demonstrates that SEP benefits our model in learning novel skills by performing skill-specific knowledge learning.

 \textbf{Ablation on memory size}. We also explore the impact of various sizes of memory buffer $\mc{M}$. As shown in Tab.~\ref{tab: replay}, with $|\mc{M}|=6$, the forgetting rate of our model is notably reduced by $1.1\%\sim10.8\%$ compared to $|\mc{M}|=4$ and $2$, respectively.  These results indicate that increasing the memory size substantially mitigates catastrophic forgetting, while introducing increased memory overhead.

 \textbf{Ablation on the external models in SSR module}. To assess the effectiveness of the SSR module’s reliance on external models, we conducted ablation studies with Base (NeRF-only rendering), Base+DINO \cite{caron2021emerging}, and Base+CLIP \cite{radford2021learning}. As shown in Tab.~\ref{tab: ablation_ssr}, using only NeRF captures the geometric structure but fails to preserve the skill-shared semantics of the 3D scene, resulting in degraded task performance. Replacing Stable Diffusion with DINO or CLIP transfers only partial semantic knowledge, yet still falls short of the full SSR module. These findings underscore the crucial role of Stable Diffusion in ensuring robust 3D skill-shared semantic consistency.

\textbf{Ablation on hyper-parameters in NBAgent}.  We analyze the influence of the trade-off parameters 
 $\lambda_1$ and $\lambda_2$  on the overall performance under the 5-5 setting. As illustrated in Fig.~\ref{fig: hyperparam}, both the ``All” and ``Avg.” metrics exhibit a similar trend, where the optimal performance is achieved when  $\lambda_1$ = 0.1 and $\lambda_2$ = 0.2. In addition, we compute the average similarity $\mc{C}_{max}$ using the data from incremental tasks under the 5+5 setting and present the intra-skill and inter-skill similarities in Fig.~\ref{fig: delta}. The results indicate that the hyperparameter $\delta = 0.8$ effectively distinguishes between old skills (with higher similarity) and new skills (with lower similarity).

\textbf{Ablation on task order}. To evaluate the effect of task order on continual learning performance, we conduct additional experiments using three different randomized task orders.  The results in Tab.~\ref{tab: order} show that NBAgent maintains stable performance across different task orders, indicating its robustness to task sequence variations. The performance fluctuations among orders are minor (within about 2\%), suggesting that catastrophic forgetting is effectively mitigated, which demonstrates that NBAgent achieves consistent NBRL regardless of the training order.

\subsection{Analysis of Task Complexity}
\label{sec: task_complexity}
As shown in Tab.~\ref{tab: longterm}, we present the comparison results between NBAgent and other methods. NBAgent consistently outperforms the baselines by $2.8\%\sim35.3\%$ in terms of the average metric. These results demonstrate that NBAgent effectively mitigates catastrophic forgetting of previously learned skills through the joint learning of skill-specific and skill-shared knowledge in long-term learning. Additionally, as shown in Fig.~\ref{fig: forgetting_rate}, we report the forgetting rates of NBAgent and the compared baselines. The forgetting rate of our approach increases more gradually than that of other methods, demonstrating its stronger resistance to catastrophic forgetting and its ability to retain learned knowledge more stably over time.

\subsection{Analysis of Computation Complexity}
\label{sec: computation_complexity}
In this subsection, we analyze the computational complexity of our framework in addressing the NBRL problem. As summarized in Tab.~\ref{tab: parameter}, we present comparative analyses with baseline methods across two critical metrics: trainable parameters and memory storage. To ensure equitable evaluation conditions, all experiments are conducted on identical hardware settings utilizing two NVIDIA A6000 GPUs. From Tab.~\ref{tab: parameter},  our framework achieves state-of-the-art parameter efficiency through the proposed SEP module. By only training the skill-specific latent space in Eq.~(\ref{eq: cross_attention}) and skill-specific LoRA layers in Eq.~(\ref{eq: lora}), the model reduces $62.6\%$ trainable parameters of the total parameters and achieve the competitive efficiency in training time per 100 iterations, while reaching the highest success rates. These results demonstrate that our model achieves an excellent optimal trade-off between computational efficiency and task effectiveness.

\section{Conclusion and Limitation}
\label{Limitation}
\textbf{Conclusion}. In this paper, we systematically investigate a pioneering Never-ending Behavior-cloning Robot Learning (NBRL) problem and propose NBAgent, a continual learning framework to continually learn skill-wise knowledge. Specifically, we design a skill-shared semantic rendering module and a skill-shared representation distillation module to tackle 3D scene representation overlooking on past encountered skills. Meanwhile, we propose a skill-specific evolving planner to decouple the skill-wise knowledge through adaptive latent space partitioning and parameter-efficient LoRA layers, enabling simultaneous effective learning of novel skills and mitigation of catastrophic forgetting. We develop two NBRL benchmarks to demonstrate the manipulation accuracy, which verifies the effectiveness of our NBAgent against baselines. Through extensive experiments on robotic manipulation tasks, we validate that NBAgent achieves state-of-the-art performance in NBRL problem.

\textbf{Limitation}. Although NBAgent achieves state-of-the-art results on the Never-ending Behavior-cloning Robot Learning (NBRL) problem, the primary limitation persists: The current implementation relies on predefined rules for keyframe extraction from demonstration datasets. While effective for constrained manipulation tasks (\emph{e.g.}, drawer opening, object grasping), these heuristics may not generalize to scenarios requiring dynamic skill segmentation, such as fluid pouring or deformable object manipulation.

\bibliographystyle{plain}
\bibliography{ref}

\end{document}